\documentclass{article}

\usepackage{arxiv}

\usepackage{multirow}
\usepackage{array}
\usepackage[ruled,vlined]{algorithm2e}

\usepackage{xcolor}

\usepackage[utf8]{inputenc} 
\usepackage[T1]{fontenc}    
\usepackage{hyperref} 
\usepackage{scalerel,stackengine}
\stackMath
\newcommand\reallywidehat[1]{%
\savestack{\tmpbox}{\stretchto{%
  \scaleto{%
    \scalerel*[\widthof{\ensuremath{#1}}]{\kern-.6pt\bigwedge\kern-.6pt}%
    {\rule[-\textheight/2]{1ex}{\textheight}}
  }{\textheight}%
}{0.5ex}}%
\stackon[1pt]{#1}{\tmpbox}%
}

\usepackage{graphicx}
\usepackage[export]{adjustbox}
\usepackage{xcolor}
\newcommand\Tstrut{\rule{0pt}{2.1ex}}       
\newcommand\Bstrut{\rule[-0.9ex]{0pt}{0pt}} 
\usepackage{subfigure}
\usepackage{colortbl}
\usepackage{paralist}
\usepackage{bbm}
\usepackage{url}            
\usepackage{booktabs}       
\usepackage{amsfonts}       
\usepackage{nicefrac}       
\usepackage{microtype}      
\usepackage{lipsum}	
\usepackage{xcolor}
\title{We Don't Speak the Same Language: Interpreting Polarization through Machine Translation}

\author{
Ashiqur R. KhudaBukhsh\thanks{Ashiqur R. KhudaBukhsh and Rupak Sarkar  are equal contribution first authors.} \\
  \small{Carnegie Mellon University} \\
  \texttt{akhudabu@cs.cmu.edu} \\
  \And
  Rupak Sarkar$^*$\\
  \small{Maulana Abul Kalam Azad University of Technology}\\
  \texttt{rupaksarkar.cs@gmail.com } \\
\And
Mark S. Kamlet \\
  \small{Carnegie Mellon University}\\
  \texttt{kamlet@cmu.edu} \\
 \And
Tom M. Mitchell \\
  \small{Carnegie Mellon University}\\
  \texttt{tom.mitchell@cs.cmu.edu} \\
}

\begin{document}
\maketitle

\begin{abstract}

Polarization among US political parties, media and elites is a widely studied topic. Prominent lines of prior research across multiple disciplines have observed and analyzed growing polarization in social media. In this paper, we present a new methodology that offers a fresh perspective on interpreting polarization through the lens of machine translation. With a novel proposition that two sub-communities are speaking in two different \emph{languages}, we demonstrate that modern machine translation methods can provide a simple yet powerful and interpretable framework to understand the differences between two (or more) large-scale social media discussion data sets at the granularity of words. Via a substantial corpus of 86.6 million comments by 6.5 million users on over 200,000 news videos hosted by YouTube channels of four prominent US news networks, we demonstrate that simple word-level and phrase-level translation pairs can reveal deep insights into the current political divide -- what is \emph{black lives matter} to one can be \emph{all lives matter} to the other.

\end{abstract}


\section{Introduction}
\emph{One mans meate is another mans poyson.}\\
-- Thomas Draxe; \emph{Bibliotheca Scholastica}; 1616.\\

\begin{figure}[t]
\centering
\includegraphics[trim={0 0 0 0},clip, height=1.8in]{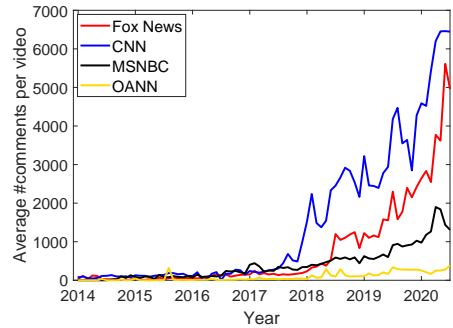}
\caption{{Temporal trend  showing number of comments made about news videos on four news networks' official YouTube channels over time.}}
\label{fig:newEngagement}
\end{figure}

Polarization among US political parties~\cite{poole1984polarization, layman2010party,mccarty2016polarized, baldwin2016past,mcconnell2017research}, media~\cite{hollander2008tuning, stroud2011niche} and elites is a widely studied topic. Studies have shown that over the last 30 years, both Democrats and Republicans have become more negative in their views toward the opposition party~\cite{iyengar2012affect}. Further, behavioral studies indicate that such negative views have affected outcomes in settings as diverse as allocating scholarship funds~\cite{iyengar2015fear}, mate selection~\cite{huber2017political}, and employment decisions~\cite{gift2015does}. Prominent lines of prior research across multiple disciplines have observed and analyzed growing polarization  in social media~\cite{demszky-etal-2019-analyzing, DBLP:journals/corr/abs-2001-02125,bakshy2015exposure,Kavanaugh2020}, and previous studies have reported substantial partisan and ideological divergence in both content and audience in major US TV news networks~\cite{NYTimeEvilTwin, bozell2004weapons, gil2012selective, hyun2016agenda}. Over the last few years, these news networks have amassed millions of subscribers in their respective YouTube channels. As a result, the overall engagement in terms of likes, views, and comments has shown a steep upward trend (see, Figure~\ref{fig:newEngagement}). Previous studies have reported news media's role in fostering partisanship~\cite{NYTimeEvilTwin,hyun2016agenda}. User engagement in YouTube news networks presents an excellent opportunity to study web-scale user behavior in response to mainstream news content. In this work, via a comprehensive analysis of a substantial corpus of 86.6 million user comments on over 200,000 YouTube videos hosted by four prominent US news networks, we present a novel approach to interpreting polarization using machine translation methods.

We ask the following question: \emph{Is it possible that the two sub-communities are speaking in two different languages such that certain words do not mean the same to the liberal and conservative viewership? If yes, how do we find those words?} \textcolor{black}{To do this in the context of user comments from the YouTube channels of different cable news networks, we begin by hypothesizing that viewers of CNN speak in what might be called ``CNN-English'' and viewers of Fox News speak in ``Fox-News-English''.   We then apply modern machine translation procedures to these two “languages” in the same way English would be translated into, say, Spanish.} 

\begin{table*}[t]

{
\scriptsize
\begin{center}
     \begin{tabular}{| p{7.5cm}  | p{7.5cm} |}
    \hline
    \Tstrut 
    \cellcolor{blue!25} Republicans are the greatest threat to America &\cellcolor{red!25} Democrats are the greatest threat to America\\
     \hline \Tstrut
\textcolor{blue}{Republicans are the greatest threat to America} that this nation has ever seen. They have willingly enabled a tyranny and wannabe dictator\ldots
 
 \Bstrut&      \ldots Had Trump placed more restrictions on travel sooner, Democrats would have cried ``racism''.  \textcolor{red}{Democrats are the greatest threat to America} today.\\ 
 \hline
    \Tstrut \cellcolor{blue!25} Republicans are traitors &\cellcolor{red!25} Democrats are traitors \\
    
    \hline
    \Tstrut
The \textcolor{blue}{Republicans are traitors}. Period, full stop. All good and patriotic Americans must see this, realize it for what it is, and then begin to act accordingly\ldots
 \Bstrut&      The \textcolor{red}{DEMOCRATS are TRAITORS} to our country and should be rounded up and exiled to a island.\\
 \hline
    \Tstrut \cellcolor{blue!25} I will never vote Republican again &\cellcolor{red!25} I will never vote Democrat again \\
     \hline \Tstrut
What a liar! I have always voted for the man not the party but after the way the republicans have acted \textcolor{blue}{I will NEVER vote republican again}. \ldots   
 
 \Bstrut&      I used to vote for the democrats because they cared about poor people.  Now they only care about exploitable non-american poor people,  talk about being un-american.  \textcolor{red}{I will never vote Democrat again.}\\
 \hline
    \Tstrut \cellcolor{blue!25} Democrats are patriots &\cellcolor{red!25} Republicans are patriots \\
     \hline \Tstrut
\textcolor{blue}{Democrats are patriots} just holding on to our constitution ! McConnell and trump must have their crowns slapped off their tyranny heads   
 
 \Bstrut&      \textcolor{red}{Republicans are patriots}.   demoRats are traitors.\\
 \hline
    \Tstrut \cellcolor{blue!25} Democrats are fighting for &\cellcolor{red!25} Republicans are fighting for \\
     \hline \Tstrut
WE ARE A NATION OF IMMIGRANTS. THAT'S WHAT MAKES AMERICA GREAT!!!\ldots DIVERSITY IS THE CORNERSTONE OF WESTERN DEMOCRACY. THE \textcolor{blue}{DEMOCRATS ARE FIGHTING FOR} EQUALITY AND ECONOMIC STABILITY\ldots
 
 & Democrats are doing everything in their power to take away your power as a citizen to make choices.  The \textcolor{red}{Republicans are fighting for} YOU as an individual.  Come on Americans!  Wake up!\ldots\\
 \hline
    \Tstrut \cellcolor{blue!25} Vote all Democrats in &\cellcolor{red!25} Vote all Republicans in \\
     \hline \Tstrut
\ldots Regardless of whether or not our candidates win in the primaries or whether we even like the Democrats we  must be prepared to \textcolor{blue}{vote all Democrats in} and all Republicans out\ldots 
 
 & We the American people are tired of these crazy dems.Hope we \textcolor{red}{vote all republicans in} office.\\
 \hline

    \end{tabular}
\end{center}
\caption{Illustrative examples highlighting that \texttt{Democrats} and \texttt{Republicans} are used in almost mirroring contexts. Left and right column contain user comments obtained from official YouTube channels of CNN and Fox news, respectively. Our translation algorithm detects $\langle\texttt{democrats}, \texttt{republicans} \rangle$ as one of many translation pairs.}
\label{tab:demrep}}
\end{table*}

\textcolor{black}{But, because both languages are using English words, the vast majority of words should translate into something very close to themselves. For instance, \texttt{grape} in CNN-English will very likely translate to \texttt{grape} or something highly similar in meaning to a \texttt{grape} in Fox-News-English, just as \texttt{tree} in Fox-News-English will very likely translate into \texttt{tree} in CNN-English or something close to \texttt{tree} in meaning.   Recognizing this, we focus on those distinct pairs of words that translate into one another but have very different meaning and usage.}  

Such pairs are not hard to envision. Consider this simple word pair: $\langle \texttt{republicans}, \texttt{democrats}\rangle$ and illustrative examples of their appearances in CNN and Fox News YouTube user discussions (listed in Table~\ref{tab:demrep}) where these two terms appear in highly similar contexts. Intuitively,  \texttt{republicans} will appear in largely favorable contexts in conservative discussion outlets while \texttt{democrats} will appear mostly in unfavorable contexts. \textcolor{black}{Conversely, in liberal discussion outlets, their roles with be completely reversed in what appear to be virtually identical contexts.}

How many such word pairs exist and what stories do they tell us? In this paper, we present a systematic approach to detect and study such word pairs developing a quantifiable framework to evaluate how \emph{similar} or \emph{dissimilar} web-scale discussions of two sub-communities are \textcolor{black}{by offering a fresh perspective on interpreting linguistic manifestation of polarization through the lens of machine translation.} With \textcolor{black}{this} novel proposition that two sub-communities are speaking in two different \emph{languages}, we demonstrate that modern machine translation methods can provide a simple yet powerful and interpretable framework to understand the differences between two (or more) large-scale social media discussion data sets at the granularity of words. 

Beyond promising results in quantifying ideological differences among multiple news networks, our automated method presents a compelling efficiency argument. It is infeasible to manually examine millions of social media posts (in the order of 100 million tokens) to identify and understand issue-centric differences. Our method boils down this task to manual inspection of less than a few hundred salient translation pairs that can provide critical insights into ideological differences. For example, translation pairs such as $\langle \texttt{solar}, \texttt{fossil}\rangle$ or $\langle \texttt{mask}, \texttt{muzzle}\rangle$ \textcolor{black}{can provide insights} to the ongoing energy debate or the debate surrounding mask and freedom of choice, and may indicate aggregate stance of a sub-community. Going beyond single-word translations, through simple phrase translations our method can reveal the current, deep political divide -- \textcolor{black}{what is \texttt{black lives matter} in CNN-English can be \texttt{all lives matter} in Fox-News-English}.

\section{Our General Idea}
A standard machine translation system that performs single word translation takes a word in a source language as input (denoted by $w_\mathit{source}$) and outputs an equivalent word in a target language (denoted by $w_\mathit{target}$). For example, in a translation system performing \emph{English} $\rightarrow$ \emph{Spanish} translation, if the input word $w_\mathit{source}$ is \texttt{hello}, the output word $w_\mathit{target}$  will be \texttt{ola}, i.e., \emph{translate} (\texttt{hello})$^{\mathit{English} \rightarrow \mathit{Spanish}}$ = \texttt{ola}. \textcolor{black}{The distributional hypothesis of words~\cite{harris1954distributional} famously stated ``\emph{You shall know a word by the company it keeps}"~\cite{firth1957synopsis}. The ``company'' of a word, i.e. the set of words that tend to occur closely to it, aka its context, plays an important role in modern machine translation methods.}  The underlying computational intuition is that in a translation pair $\langle w_\mathit{source}, w_\mathit{target}  \rangle$, the contexts in which $w_\mathit{source}$ appears in the source language are highly similar with the contexts in which $w_\mathit{target}$ appears in the target language.


A powerful way to operationalize the notion of words being close (or far) from one another is to employ a method which embeds each word as a vector in a high-dimensional space (referred to as an embedding) and using the proximity of any two words in that space as a measure of closeness. This approach, set forth in~\cite{mikolov2013distributed}, initiated a rich line of machine translation literature. 
\cite{mikolov2013exploiting} first observed that continuous word embedding spaces exhibit similar structures across languages and proposed a linear mapping from a source to target embedding space. Their approach worked surprisingly well even in distant language pairs. Since then, several studies proposed improvements over this general idea of learning cross-lingual embedding spaces~\cite{faruqui2014improving, xing2015normalized, lazaridou2015hubness, ammar2016massively}.


In our work, we are interested in leveraging this machine translation literature to user discussions taking place at the comments section of official YouTube channels of two different news networks (e.g., CNN and Fox News). \textcolor{black}{As we already mentioned}, of course, both the CNN and Fox News corpora are in English.  But we introduce a novel and powerful approach by treating them as two different languages: $\mathcal{L}_\mathit{cnn}$ and $\mathcal{L}_\mathit{fox}$.  Given that our ``languages'' are actually English from different sub-communities, on most occasions, $\langle w_\mathit{source}$, $w_\mathit{target}\rangle$ will be identical word pairs (e.g., $\langle \texttt{grape}, \texttt{grape} \rangle$); i.e., for a given translation direction (say, ${\mathcal{L}_\mathit{cnn} \rightarrow \mathcal{L}_\mathit{fox}}$), \emph{translate} ($w_\mathit{source}$)$^{\mathcal{L}_\mathit{cnn} \rightarrow \mathcal{L}_\mathit{fox}}$ = $w_\mathit{source}$. The interesting cases are the pairs that include two different English words, i.e., $\langle w_\mathit{source}$, $w_\mathit{target}\rangle$ such that $ w_\mathit{source} \ne w_\mathit{target}$. We call such word pairs \emph{misaligned pairs}. 

These different word pairs can arise for either of two very different phenomena, though both result in the same treatment by our translation algorithm, because both phenomena result in the fact that  $w_\mathit{source}$ is used by one sub-community in very similar contexts  $w_\mathit{target}$ is used by the other sub-community.

\textcolor{black}{One case of misaligned pairs} is where both $w_\mathit{source}$ and $w_\mathit{target}$ in the pair $\langle w_\mathit{source},w_\mathit{target} \rangle$ refer to the \textcolor{black}{actual same grounded entity} (e.g., $\langle \texttt{pelosi}, \texttt{pelousy} \rangle$). \textcolor{black}{So, for instance in ``Pelosi spoke yesterday'' and ``Pelousy spoke yesterday'', both communities are referring to Nancy Pelosi, the speaker of the United States House of Representatives.}  In this case, the reason that the two words appear in the same context is that the two communities are stating very similar beliefs about that entity. In this case, we can think of the two words as synonyms referring to the same entity, though the difference in the actual names can reflect important differences in attitudes toward that entity.  The second case is where the word pair refers to two different entities, as in $\langle \texttt{tapper},\texttt{hannity} \rangle$.  Here, the phenomenon detected is that one sub-community makes statements about $w_\mathit{source}$ that are very similar to the statements made by the second sub-community about word $w_\mathit{target}$ (e.g., ``Tapper is a great interviewer'' vs. ``Hannity is a great interviewer'').  Table~\ref{tab:pairsPreview} characterizes additional phenomena that can produce word pairs, though each of the rows there correspond to one of the two phenomena above.  That is, the examples under political entities, news entities, and ideological rows in Table~\ref{tab:pairsPreview} correspond to word pairs that refer to different entities. The examples under the derogatory, synonyms, and spelling errors rows correspond to the case where the two words refer to the same entity.

\begin{table}[t]

{
\scriptsize
\begin{center}
     \begin{tabular}{| p{1.8cm}  |  p{5.7cm} |}
    \hline
    \Tstrut Category  & Misaligned pairs \\
     \hline 
 Political entities & 
 $\langle \texttt{democrats}, \texttt{republicans} \rangle$, $\langle\texttt{nunes},\texttt{schiff}\rangle$ \\
    \hline 

 News entities &  $\langle \texttt{fox}, \texttt{cnn} \rangle$, $\langle \texttt{tapper}, \texttt{hannity} \rangle$     \\
    \hline 
    	
 Derogatory & 
 $\langle \texttt{chump}, \texttt{trump} \rangle$, 
 $\langle \texttt{pelosi}, \texttt{pelousy} \rangle$
     \\
    \hline
(Near) synonyms & 
$\langle \texttt{lmao}, \texttt{lol} \rangle$, 
$\langle \texttt{allegations}, \texttt{accusations} \rangle$
\\
    \hline

Spelling errors & 
$\langle \texttt{mueller}, \texttt{muller} \rangle$,  
$\langle \texttt{hillary}, \texttt{hilary} \rangle$ 
     \\
    \hline

Ideological & 
$\langle \texttt{kkk}, \texttt{blm} \rangle$ 
$\langle \texttt{liberals}, \texttt{conservatives} \rangle$      \\
    \hline

    \end{tabular}
\end{center}
\vspace{0.5cm}
\caption{{Examples of misaligned word pairs. Word pairs are presented in $\langle w_\mathit{cnn}, w_\mathit{fox} \rangle$ format where $w_\mathit{cnn} \in \mathcal{L}_\mathit{cnn}$ and $w_\mathit{fox} \in \mathcal{L}_\mathit{fox}$. A detailed treatment with more examples is presented in Section~\ref{sec:results}.}}
\label{tab:pairsPreview}}
\end{table}


Our intuition is misaligned pairs may reveal useful insights into differences between the two sub-communities. For example, \texttt{solar} in $\mathcal{L}_\mathit{cnn}$ translating into \texttt{fossil} in  $\mathcal{L}_\mathit{fox}$ possibly indicates that the two communities have divergent, and close to mirror image views of climate change and renewable energy.  Or, \texttt{cooper} in $\mathcal{L}_\mathit{cnn}$ translating into \texttt{hannity} in  $\mathcal{L}_\mathit{fox}$ possibly indicates that the CNN sub-community views Anderson Cooper favorably and Sean Hannity unfavorably while the Fox News sub-community views the two news entities exactly the opposite way.  


\noindent\textbf{Quantifying similarity and dissimilarity:} If two sub-communities use most words in similar contexts, the number of misaligned word pairs will be fewer than the number of misaligned word pairs if the two communities use a large number of words (e.g., entities, issues) in different contexts. We can thus construct a measure of similarity and dissimilarity between discussions in sub-communities by computing the fraction of misaligned words over the size of the source vocabulary -- the larger this number, the greater the dissimilarity. Comparing across multiple corpora will require careful selection of source and target vocabulary and several other design decisions to ensure cross-corpus comparability. A formal treatment of our approach is presented in Section~\ref{sec:framework}.

\section{Data Set}
Our data set consists of user comments posted on videos hosted by four US news networks' official YouTube channels listed in Table~\ref{tab:channels}. CNN, Fox News, and MSNBC are considered to be the three leading cable news networks in the US~\cite{NewsNetworkRank}. Commensurate to their cable TV popularity, these three channels have a strong YouTube presence with millions of subscribers. Our choice of OANN, a conservative media outlet, is guided by the observation that the current US President shares favorable views about this network on social media platforms~\cite{OANTrump}.


\begin{table}[b]
{
\scriptsize
\begin{center}
     
\begin{tabular}{|l | c | c |}
    \hline
    News Network & \#Subscribers & \#Videos \\
    \hline                                 
   CNN (Cable News Network)  &  10.6M & 95,433 \\
    \hline
   Fox News & 6.09M & 65,337 \\
    \hline
  MSNBC  &  3.45M & 31,732 \\
    \hline
  OANN (One America News Network) &  0.84M & 11,884 \\
    \hline
    \end{tabular}

\vspace{0.2cm}
\caption{List of news networks considered. Video count reflects \#videos uploaded on or before 31 July 2020 starting from 1 January 2014.}
\label{tab:channels}
\end{center}
}
\end{table}

Starting from 1 January, 2014, we considered videos uploaded on or before 31 July, 2020. We used the publicly available YouTube API to collect comments from these videos. YouTube comments exhibit a two-level hierarchy. Top-level comments can be posted in response to a video and
replies can be posted to these top-level comments. We collect both and for the
analyses in this paper, we focus on the top-level comments. Analyses on replies are presented in the Appendix. Overall, we obtain 86,610,914 million comments (50,988,781 comments and 35,622,133 replies) on 204,386 videos posted by 6,461,309 unique users.  We use standard preprocessing (e.g., punctuation removal, lowercasing) for our comments. The preprocessing steps are described in details in the Appendix.  

In what follows, we present a few notable results we obtain while analyzing user engagement. The Appendix contains additional results.   

\subsection{Temporal Trends of Video Likes and Dislikes:}

\begin{figure}[htb]
\centering
\includegraphics[trim={0 0 0 0},clip, height=1.8in]{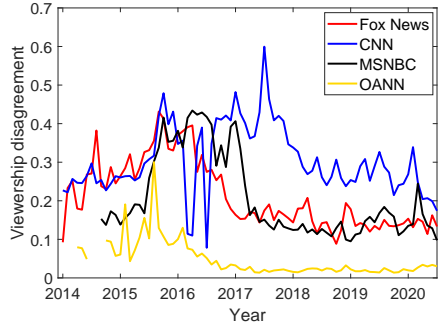}
\caption{{Temporal trend of  viewership disagreement in terms of video likes and dislikes. Each point in the graph represents the monthly average of $\frac{v_\mathit{dislike}}{v_\mathit{like} + v_\mathit{dislike}}$} for each video $v$ uploaded on the news network's official YouTube channel. We report this value only if 10 or more videos are uploaded in a given month for a specific channel.} 
\label{fig:likeEvolution}
\end{figure}

We first introduce a simple measure to evaluate viewership disagreement. For a given video $v$, let $v_\mathit{like}$ and $v_\mathit{dislike}$ denote the total number of likes and dislikes $v$ received. For each video $v$, we first compute the ratio $\frac{v_\mathit{dislike}}{v_\mathit{like} + v_\mathit{dislike}}$. If a video is disliked by fewer viewers and liked by a large number of viewers, this value will be close to 0. Conversely, if the video is overwhelmingly disliked, the value will be close to 1. A value close to 0.5 indicates that the opinion about the video among the viewership is divided.  Formally, let $\mathbbm{I}(v, m)$ be an indicator function that outputs 1 if video $v$ is uploaded in month $m$ and outputs 0 otherwise. For a given channel and a particular month $m^j$, we compute the following viewership disagreement factor: \large{$\frac{\Sigma_i \mathbbm{I}(v^i, m^j) \frac{v^i_\mathit{dislike}}{v^i_\mathit{dislike} + v^i_\mathit{like}}}{\Sigma_i \mathbbm{I}(v^i, m^j)}$}.\\
\normalsize
Our disagreement measure has the following advantages. First, assuming $v_\mathit{like} + v_\mathit{dislike} \ne 0$,  $0 \le \frac{v_\mathit{dislike}}{v_\mathit{like} + v_\mathit{dislike}} \le 1$. Since average of bounded variables is also bounded, our disagreement factor is also bounded within the same range [0, 1]. Since each video's disagreement measure is bounded within the range [0, 1], this measure is robust to outliers; a single heavily liked or disliked video cannot influence the overall average by more than $\frac{1}{n}$ where $n$ is the total number of videos uploaded in that particular month\footnote{Figure~\ref{fig:likeEvolution} presents the temporal trend of the viewership disagreement factor for four major news networks during the time period of 2014-2020. More than 100 videos are uploaded for most of the months we considered in our analysis shown in Figure~\ref{fig:likeEvolution}.}.

Figure~\ref{fig:likeEvolution} presents the temporal trend of the viewership disagreement factor for four major news networks during the time period of 2014-2020. A paired t-test reveals that beyond 2017, the viewership disagreement among CNN viewers is larger than all other channels' viewership disagreement with p-value less than 0.0001. This indicates that possibly, CNN is less of an echo chamber as compared to the other three media outlets. Among the four news networks we considered, the viewership disagreement among OANN viewers is the lowest. Our results corroborate to previous findings on the existence of echo chambers in highly conservative social media platforms~\cite{zannettou2018gab, horne2019different}. While not presented as a formal study, a parallel between Fox News and MSNBC's comparable partisanship, albeit for two different political views, has been reported before~\cite{NYTimeEvilTwin}. 
Adding evidence to this observation, in Figure~\ref{fig:likeEvolution}, we note that the temporal trends of viewership disagreement is similar across MSNBC and Fox News.

\section{Related Work}

Polarization and partisanship in US politics is a widely studied topic with surveys and studies focusing on diverse aspects such as congressional votes~\cite{poole1984polarization}, response to climate change~\cite{fisher2013does,baldwin2016past}, polarization in media~\cite{prior2013media} and economic decisions~\cite{mcconnell2017research}, and partisanship in search behavior~\cite{krupenkin2019president} to name a few. Our work contrasts with recent  computational social science research on polarization~\cite{demszky-etal-2019-analyzing,Kavanaugh2020} along the following three main dimensions: (1) our fresh perspective on casting the task of quantifying polarization as a machine translation problem; (2) our broad treatment of the problem without focusing on specific events or type of events; and (3) our focus on YouTube data of major news networks. Unlike~\cite{demszky-etal-2019-analyzing} that focused on a specific type of events (mass-shootings) and~\cite{Kavanaugh2020} that studied controversy surrounding the most-recent supreme court confirmation, we consider a longer, continuous time-horizon (2017 - 2020) within which two (or more) sub-communities discuss a broad range of issues. Priors lines of research on quantifying polarization among YouTube and Facebook users have focused on characterizing user behavior in the context of scientific and conspiratorial content consumption~\cite{bessi2016users}. 

While work focusing on dialectical variants of English and their detection challenges exists~\cite{blodgett-etal-2016-demographic,eisenstein2011discovering}, to the best of our knowledge, treating two US news networks' discussions as two distinct languages and leveraging modern machine translation literature~\cite{SmithTHH17} to detect mismatched translation pairs has not been explored before. 

A recent work has focused on 38 prominent Indian news networks' YouTube channels leading up to 100 days of 2019 Indian General election and has used language models to mine insights~\cite{election}.  This study reported evidence of religious polarization in India. Our work addresses the challenge of quantifying intra-news network user discussion differences using a novel approach of machine translation. 

Presence of human biases in word embedding in social media corpora is a well-established observation~\cite{caliskan2017semantics,garg2018word}. Recent lines of work channelised considerable efforts to debias such embedding~\cite{bolukbasi2016man,manzini-etal-2019-black}. Our work presents a novel method to detect word pairs where two different sub-communities exhibit comparable biases for two different words across two different corpora.

\section{Framework and Design Choices}\label{sec:framework}
We first describe our framework for two news networks' discussion data sets we assume to be authored in two different \emph{languages}: $\mathcal{L}_\mathit{s}$ and $\mathcal{L}_\mathit{t}$. A more general treatment involving more than two news networks is presented in Section~\ref{sec:twoOrMore}. Let $\mathcal{D}_\mathit{s}$ and $\mathcal{D}_\mathit{t}$ be two \emph{monolingual} text corpora authored in languages $\mathcal{L}_\mathit{s}$ and $\mathcal{L}_\mathit{t}$, respectively. Let with respect to the corpora $\mathcal{D}_\mathit{s}$ and $\mathcal{D}_\mathit{t}$,  $\mathcal{V}_\mathit{s}$ and $\mathcal{V}_\mathit{t}$ denote the source and target vocabularies, respectively. Let $w^{e,l}$ denote the vector representation of the word $w$ in an embedding space trained on $\mathcal{D}_\mathit{l}$. 
A word translation scheme, $\mathcal{L}_{s} \rightarrow \mathcal{L}_{t}$, takes a word $w_{source} \in \mathcal{V}_{s}$ as input and outputs a single word translation $w_{target}$ such that \\ (1) $w_{target} \in \mathcal{V}_{t}$, and \\ (2) $\forall w \in \mathcal{V}_{t}$, \emph{dist}($w_{source}^{e,s} W$, $w^{e,t}$) $\ge$ \emph{dist}($w_{source}^{e, s} W$, $w_{target}^{e,t}$) where $W$ is a transformation matrix.

\subsection{Design Choices}

We now describe and justify our design choices.

\subsubsection{Translation algorithm} We compute $W$ using a well-known algorithm ~\cite{SmithTHH17}. This algorithm requires two monolingual corpora and a bilingual seed lexicon of word translation pairs as inputs. First, two separate monolingual word embedding are induced using a monolingual word embedding learning model. Following~\cite{SmithTHH17}, we use FastText~\cite{bojanowski2017enriching} to train monolingual embedding. Next, the bilingual seed lexicon is used to learn an orthogonal transformation matrix, which is then used to align the two vector spaces. Finally, to translate a word from the source language to the target language, we multiply the embedding of the source word with the transformation matrix to align it with the target vector space. Then, the nearest neighbour of the aligned word vector in the target vector space is selected as the translation of the source word in the target language. Following~\cite{SmithTHH17}, we use cosine distance as our distance metric. Our choice of the translation algorithm is motivated by its (1) competitive performance~\cite{SmithTHH17}, (2) simple and elegant design, and (3) robustness to lexicon sparsity. The Appendix contains qualitatively similar results with a different translation algorithm~\cite{LampleCRDJ18}.

Unlike typical machine translation task, we are dealing with two English corpora. For the seed lexicon of translation pairs, ideally, we would prefer words that are neutral across political beliefs with very high probability. To this end, we construct our seed lexicon with English stopwords in the following way: \{$\langle w, w \rangle$\}, $w$ is a stopword. We consider the default English stopword set of NLTK~\cite{nltk}. 

\subsubsection{$\mathcal{V}_\mathit{s}$ and $\mathcal{V}_\mathit{t}$} We first ensure both corpora have identical size (in terms of \#tokens). We next concatenate token-balanced $\mathcal{D}_\mathit{s}$ and $\mathcal{D}_\mathit{t}$ and choose the top 5,000 and top 10,000 words by frequency in the combined corpus as the source vocabulary $\mathcal{V}_\mathit{s}$ and target vocabulary $\mathcal{V}_\mathit{t}$, respectively. We exclude the stopwords while computing $\mathcal{V}_\mathit{s}$ and $\mathcal{V}_\mathit{t}$ since we use them as anchor words for our translation algorithm. Note that, here we slightly abuse the notation since we have identical $\mathcal{V}_\mathit{s}$ and $\mathcal{V}_\mathit{t}$ across both translation directions.

We token-balance our corpora to (1) enforce that the quality of the embedding is comparable across corpora and (2) ensure that both corpora have a fair influence on words that are included in $\mathcal{V}_\mathit{source}$ and $\mathcal{V}_\mathit{target}$.

\subsubsection{\emph{Similarity} ($\mathcal{L_\mathit{s}}, \mathcal{L_\mathit{t}})$} The similarity measure between two languages along a given translation direction computes the fraction of words in $\mathcal{V}_\mathit{s}$  that translates to itself, i.e.,\\
\large{
\emph{Similarity}($\mathcal{L_\mathit{s}}, \mathcal{L_\mathit{t}})$ = $\frac{\Sigma_{w \in \mathcal{V}_\mathit{s}} \mathbbm{I}(\emph{translate}(w)^{\mathcal{L}_\mathit{s} \rightarrow \mathcal{L}_\mathit{t}} = w)}{|\mathcal{V}_\mathit{s}|}$
}.
\normalsize
The indicator function returns 1 if the word translates to itself and 0 otherwise. The larger the value of \emph{Similarity} ($\mathcal{L_\mathit{s}}, \mathcal{L_\mathit{t}})$, the greater is the similarity between a language pair.

\subsubsection{Assigning user to a specific channel} It is not possible to unambiguously identify if a YouTube user prefers CNN over Fox News or not. We assign a user to CNN or Fox News using a simple filter. If a user has commented more on Fox News videos than on CNN videos during our period of interest\footnote{All our analyses are performed at the temporal granularity of a year except for 2020, where we consider the time period starting from January 1, 2020 to July 31, 2020.}, we assign her to Fox News and vice versa. While computing the discussion data set for a specific channel, we restrict ourselves to users assigned to that channel. We acknowledge that our filter makes certain assumptions that may not hold in the wild. It is possible that a user only comments on a video if she does not agree with its content. The Appendix contains additional results showing the qualitative nature of our analyses remain unchanged with or without this filter.

\subsection{Extending to Multiple News Networks}\label{sec:twoOrMore}

It is straight-forward to extend our method to more than two news network discussions data sets. We assume each discussion data set is authored in a distinct \emph{language} (CNN: $\mathcal{L}_\mathit{cnn}$; Fox News:  $\mathcal{L}_\mathit{fox}$; MSNBC:  $\mathcal{L}_\mathit{msnbc}$; and OANN: $\mathcal{L}_\mathit{oann}$). We  token-balance all corpora to identical size. Next, we concatenate all corpora and compute $\mathcal{V}_\mathit{s}$ and $\mathcal{V}_\mathit{t}$ as the top 5,000 and 10,000 words by frequency, respectively. Finally, for each translation direction, we compute pairwise similarity.  
\section{Results}\label{sec:results}
We first focus on Fox News and CNN, the two most popular news networks, and present a  qualitative analysis of the misaligned pairs obtained in the years 2017, 2018, 2019 and 2020.

\noindent\textbf{Characterizing the misaligned pairs:} Upon manual inspection, we identify the following high-level categories in the misaligned pairs listed in Table~\ref{tab:pairs}. Note that, we do not intend these categories
to be formal or exhaustive, but rather to be illustrative of the types of misaligned pairs we encountered. Further, we realize that the misaligned pairs have the following nuance. Some of the pairs map to the same entity (e.g., $\langle \texttt{liberals}, \texttt{libtards} \rangle$), while the rest map to completely different entities and beliefs (e.g., $\langle \texttt{nunes}, \texttt{schiff}\rangle$, $\langle\texttt{socialism}, \texttt{capitalism}\rangle$). 


\begin{table}[htb]

{
\scriptsize
\begin{center}
     \begin{tabular}{| p{1.8cm}  |  p{5.7cm} |}
    \hline
    \Tstrut Category  & Misaligned pairs \\
     \hline 
 Political entities & 
 $\langle \texttt{democrats}, \texttt{republicans} \rangle$, $\langle\texttt{nunes}, \texttt{schiff} \rangle$, $\langle \texttt{dem}, \texttt{republican} \rangle$, $\langle \texttt{dnc}, \texttt{gop} \rangle$,
  
 $\langle \texttt{kushner}, \texttt{burisma} \rangle$, 
 $\langle \texttt{gop},	\texttt{democrats}\rangle$,
 $\langle \texttt{flynn},	\texttt{hillary}\rangle$\\
    \hline 

 News entities &  $\langle \texttt{fox}, \texttt{cnn} \rangle$, $\langle \texttt{hannity}, \texttt{cuomo} \rangle$, $\langle \texttt{tapper}, \texttt{hannity} \rangle$, $\langle \texttt{tucker}, \texttt{cuomo} \rangle$      \\
    \hline 
    	
 Derogatory & 
 $\langle \texttt{trumptards}, \texttt{snowflakes} \rangle$,  $\langle \texttt{chump}, \texttt{trump} \rangle$, $\langle \texttt{liberals}, \texttt{libtards} \rangle$, 
 $\langle \texttt{pelosi}, \texttt{pelousy} \rangle$, 
 $\langle \texttt{obamas}, \texttt{obummer} \rangle$,  $\langle \texttt{cooper}, \texttt{giraffe} \rangle$, 
 $\langle \texttt{biden}, \texttt{creep} \rangle$,  $\langle \texttt{schiff}, \texttt{schitt} \rangle$, 
 $\langle \texttt{barr}, \texttt{weasel} \rangle$
     \\
    \hline
(Near) synonyms & 
$\langle \texttt{lmao}, \texttt{lol} \rangle$, 
$\langle \texttt{allegations}, \texttt{accusations} \rangle$, $\langle \texttt{puppet}, \texttt{stooge} \rangle$, $\langle \texttt{bs}, \texttt{bullshit} \rangle$, 
$\langle \texttt{potus}, \texttt{president} \rangle$, 
$\langle \texttt{hahaha}, \texttt{lol} \rangle$  
\\
    \hline

Spelling errors & 
$\langle \texttt{mueller}, \texttt{muller} \rangle$, $\langle \texttt{kavanaugh}, \texttt{cavanaugh} \rangle$,  
$\langle \texttt{hillary}, \texttt{hilary} \rangle$, 
$\langle \texttt{isreal}, \texttt{israel} \rangle$ 
     \\
    \hline

Ideological & 
$\langle \texttt{kkk}, \texttt{blm} \rangle$, $\langle \texttt{christianity}, \texttt{multiculturalism} \rangle$, 
$\langle \texttt{sham}, \texttt{impeachment} \rangle$, $\langle \texttt{antifa}, \texttt{nazi} \rangle$, 
$\langle \texttt{liberals}, \texttt{conservatives} \rangle$, $\langle \texttt{communism}, \texttt{nazism} \rangle$, 
$\langle \texttt{leftists}, \texttt{fascists} \rangle$, 
$\langle \texttt{liberalism}, \texttt{conservatism} \rangle$, $\langle \texttt{communists}, \texttt{nazis} \rangle$, 
 $\langle \texttt{immigrants}, \texttt{illegals} \rangle$       \\
    \hline

    \end{tabular}
\end{center}
\vspace{0.5cm}
\caption{{Characterizing the misaligned word pairs. We consider Fox News and CNN user discussions for the years 2017, 2018, 2019, and 2020. Word pairs are presented in $\langle w_\mathit{cnn}, w_\mathit{fox} \rangle$ format where $w_\mathit{cnn} \in \mathcal{L}_\mathit{cnn}$ and $w_\mathit{fox} \in \mathcal{L}_\mathit{fox}$.}}
\label{tab:pairs}}
\end{table}

We notice several misaligned pairs between political oppositions (e.g., $\langle \texttt{democrats},\texttt{republicans} \rangle$) and news entities ($\langle \texttt{fox}, \texttt{cnn} \rangle$). This result was not surprising as we have already seen in Table~\ref{tab:demrep} that Republicans and Democrats are used in almost interchangeable contexts across the two news networks' user discussions. Similarly, a CNN viewer is likely to have favorable opinion toward CNN and their anchors while a Fox viewer will have positive views toward Fox News entities (see Appendix for more details). 

Along with a few instances of (near)-synonyms\footnote{23\% out of a randomly sampled 100 misaligned pairs.} and incorrect spellings\footnote{3\% out of a randomly sampled 100 misaligned pairs.} present in some of our misaligned pairs,
we notice several derogatory terms for political and news entities (e.g., $\langle \texttt{obamas}, \texttt{obummer} \rangle$ or $\langle \texttt{chump}, \texttt{trump} \rangle$). Some of these derogatory terms could be possibly influenced by prominent public figures openly using them (e.g., $\langle \texttt{schiff}, \texttt{schitt} \rangle$)~\cite{AdamSchiff}. The derogatory terms used to describe opposition party's fervent supporters (e.g., $\langle \texttt{trumptards}, \texttt{slowflakes} \rangle$) also translate to each other.  

We observe hints of the longstanding racial debate in some of the misaligned pairs (e.g., $\langle \texttt{kkk}, \texttt{blm}\rangle$, $\langle \texttt{white}, \texttt{black}\rangle$). In Table~\ref{tab:blmkkk}, we list illustrative examples of their appearances in CNN and Fox News user discussions where these two terms appear in highly similar contexts. 

\begin{table*}[htb]

{
\scriptsize
\begin{center}
     \begin{tabular}{| p{7.5cm}  | p{7.5cm} |}
    \hline
    \Tstrut \cellcolor{blue!25} KKK is a hate group &\cellcolor{red!25} BLM is a hate group \\
     \hline \Tstrut
\ldots The \textcolor{blue}{kkk is a hate group}. But drump will not call them that, he calls them very fine people\ldots
 
 \Bstrut
 &      \ldots \textcolor{red}{blm is a hate group}.  A group of black supremacy isn't any different than white supremacy.  Defund the department of education.\\ 
 \hline
    \Tstrut 
    \cellcolor{blue!25} KKK terrorists &\cellcolor{red!25} BLM terrorists \\
    
    \hline
    \Tstrut
REPUBLICANS HAVE ALWAYS BEEN NEO-NAZI'S AND \textcolor{blue}{KKK TERRORISTS} \Bstrut
 &    Step 1 - Leftist defund the police\newline
 Step 2 - Antifa and \textcolor{red}{BLM terrorists}, looters and rioters invade neighborhoods Step 3 - Patriots (thanks to the 2nd amendment) respond to defend their families and light up the terrorists\newline Step 4 - Anitfa and BLM call the police for help and get no answer, repeat step 3 as needed\\ 
 \hline
    \Tstrut 
    \cellcolor{blue!25} KKK is nothing more than a &\cellcolor{red!25} BLM is nothing more than a  \\
     \hline \Tstrut
\textcolor{blue}{kkk is nothing more than a} low-life  racist terrorist gang\ldots 
 \Bstrut
 &      \textcolor{red}{BLM is nothing more than a} racist cult.\\
 \hline

    \end{tabular}
\end{center}
\caption{Illustrative examples highlighting that the discovered misaligned pair $\langle \texttt{blm}, \texttt{kkk} \rangle $ are used in almost mirroring contexts. Left and right column contain user comments obtained from CNN and Fox news, respectively. }
\label{tab:blmkkk}}
\end{table*}


\subsection{Comparing Multiple News Networks} 

We now perform a quantitative analysis between CNN, MSNBC and Fox News. Table~\ref{tab:threeChannels} presents the pairwise similarity between $\mathcal{L}_\mathit{cnn}$, $\mathcal{L}_\mathit{fox}$, and $\mathcal{L}_\mathit{msnbc}$. We first note that our similarity measure is reasonably symmetric; $\emph{Similarity} (\mathcal{L}_i, \mathcal{L}_j)$ and $\emph{Similarity} (\mathcal{L}_j, \mathcal{L}_i)$ have comparable values across all $i, j$. We next note that $\mathcal{L}_\mathit{msnbc}$ is more similar to $\mathcal{L}_\mathit{cnn}$ than $\mathcal{L}_\mathit{fox}$. $\mathcal{L}_\mathit{cnn}$ is more similar to $\mathcal{L}_\mathit{msnbc}$ than $\mathcal{L}_\mathit{fox}$, and $\mathcal{L}_\mathit{fox}$ is least similar to $\mathcal{L}_\mathit{msnbc}$. Hence, depending on the user discussions in their respective official YouTube channels, if we seek to arrange these three news networks along a political spectrum, a consistent arrangement is the following: MSNBC, CNN and Fox News (from left to right).      

\begin{table}[b]
\centering
\scriptsize
\setlength{\extrarowheight}{2pt}
\begin{tabular}{cc|c|c|c|}
  & \multicolumn{1}{c}{} & \multicolumn{3}{c}{$\mathcal{L}_{\emph{target}}$} \\
  & \multicolumn{1}{c}{} & \multicolumn{1}{c}{$\mathcal{L}_{\emph{cnn}}$}  & \multicolumn{1}{c}{$\mathcal{L}_{\emph{fox}}$}  & \multicolumn{1}{c}{$\mathcal{L}_{\emph{msnbc}}$} \\\cline{3-5}
            & $\mathcal{L}_{\emph{cnn}}$ &\cellcolor{blue!25} - & \textcolor{black}{90.20\%}  & \textcolor{black}{94.20\%}
 \\ \cline{3-5}
$\mathcal{L}_{\emph{source}}$  & $\mathcal{L}_{\emph{fox}}$ & \textcolor{black}{89.60\%} &\cellcolor{blue!25} - & \textcolor{black}{88.70\%}
 \\\cline{3-5}
            & $\mathcal{L}_{\emph{msnbc}}$ & \textcolor{black}{94.10\%} & \textcolor{black}{88.50\%}  &\cellcolor{blue!25} - \\\cline{3-5}
\end{tabular}
\vspace{0.5cm}
\caption{{Pairwise similarity between languages computed for the year 2020. Each corpus has identical number of tokens. The evaluation set (5K words) is computed by concatenating all three corpora and taking the top 5K words ranked by frequency. Since stopwords are used as anchor words, stopwords are excluded in the evaluation set. Appendix contains qualitatively similar results for other years. Appendix also contains  results that use a different similarity measure which  considers neighborhoods of the source and target words in their respective embedding spaces.}}
\label{tab:threeChannels}
\end{table}

One may wonder if synonymous words are causing this  perception that $\mathcal{L}_\mathit{cnn}$ and $\mathcal{L}_\mathit{msnbc}$  are closer than $\mathcal{L}_\mathit{cnn}$ and $\mathcal{L}_\mathit{fox}$. We manually examine all misaligned pairs along the translation direction where $\mathcal{L}_{cnn}$ is the source. We found that even after manually removing the synonymous misaligned pairs, our conclusion still holds. Table~\ref{tab:msnbcfox} lists a random sample of unique misaligned pairs obtained along $\mathcal{L}_\mathit{msnbc} \rightarrow \mathcal{L}_\mathit{fox}$ and $\mathcal{L}_\mathit{msnbc} \rightarrow \mathcal{L}_\mathit{cnn}$ translation directions.

\begin{table}[t]

{
\scriptsize
\begin{center}
     \begin{tabular}{| p{3cm}  | p{3cm} |}
    \hline
    \Tstrut $\mathcal{L}_\mathit{cnn} \rightarrow \mathcal{L}_\mathit{fox}$  & $\mathcal{L}_\mathit{cnn} \rightarrow \mathcal{L}_\mathit{msnbc}$ \\
     \hline 
 $\langle \texttt{trumpty}, \texttt{obummer} \rangle$& $\langle \texttt{dumpty}, \texttt{trumpty} \rangle$
        \\
    \hline
 $\langle \texttt{white}, \texttt{black} \rangle$& $\langle \texttt{nationalist}, \texttt{nazi} \rangle$
        \\
    \hline
 $\langle \texttt{pence}, \texttt{biden} \rangle$& $\langle \texttt{anderson}, \texttt{rachel} \rangle$
        \\
    \hline
 $\langle \texttt{supremacist}, \texttt{radical} \rangle$& $\langle \texttt{demonrats}, \texttt{demoncrats} \rangle$
        \\
    \hline
$\langle \texttt{socialist}, \texttt{capitalist} \rangle$& $\langle \texttt{scientist}, \texttt{expert} \rangle$
        \\
    \hline    

    \end{tabular}
\end{center}
\vspace{0.5cm}
\caption{{Discovered misaligned pairs from CNN to Fox News and MSNBC. For a translation direction $\mathcal{L}_i\rightarrow\mathcal{L}_j$, we present word pairs in $\langle w_\mathit{i}, w_\mathit{j} \rangle$ format where $w_\mathit{i} \in \mathcal{L}_\mathit{i}$ and $w_\mathit{j} \in \mathcal{L}_\mathit{j}$.}}
\label{tab:msnbcfox}}

\end{table}

\subsection{Primetime Comedies}

\begin{table}[htb]
\centering
\scriptsize
\setlength{\extrarowheight}{2pt}
\begin{tabular}{cc|c|c|c|c|c|c|c|c|c|c|c|c|c|}
  & \multicolumn{1}{c}{} & \multicolumn{4}{c}{$\mathcal{L}_\mathit{target}$} \\
  & \multicolumn{1}{c}{} &
  \multicolumn{1}{c}{$\mathcal{L}_\mathit{cnn}$}
  &\multicolumn{1}{c}{$\mathcal{L}_\mathit{fox}$}  & \multicolumn{1}{c}{$\mathcal{L}_\mathit{msnbc}$}  & \multicolumn{1}{c}{$\mathcal{L}_\mathit{comedy}$} 
  \\\cline{3-6}
          \multirow{4}{*}{$\mathcal{L}_\mathit{source}$}    & $\mathcal{L}_\mathit{cnn}$ & \cellcolor{blue!25} - &  88.7\% & 90.3\% & 83.2\%  \\ \cline{3-6}
            & $\mathcal{L}_\mathit{fox}$ & 88.7\% & \cellcolor{blue!25}- & 85.7\% & 75.0\%  \\ \cline{3-6}
            & $\mathcal{L}_\mathit{msnbc}$ & 90.3\% & 85.8\% & \cellcolor{blue!25}- & 78.4\% \\ \cline{3-6}
            & $\mathcal{L}_\mathit{comedy}$ & 81.9\% & 74.6\% & 78.0\% & \cellcolor{blue!25}-  \\ \cline{3-6}
     
\end{tabular}
\vspace{0.5cm}
\caption{{Pairwise similarity between languages computed for the year 2019. Each corpus has identical number of tokens. The evaluation set (5K words) is computed by concatenating all three corpora and taking the top 5K words ranked by frequency.}}
\label{tab:comedy}
\end{table}

\noindent\emph{Which language do viewers of prime time comedies speak?} We construct a data set of 4,099,081 comments (the Appendix contains more details about this data set) from official YouTube channels of well-known comedians focusing on political comedies (Trevor Noah, Seth Meyers, Stephen Colbert, Jimmy Kimmel, and John Oliver). Table~\ref{tab:comedy} shows that the language of YouTube primetime comedy consumers, $\mathcal{L}_\mathit{comedy}$, is farthest from $\mathcal{L}_\mathit{fox}$ and closest to $\mathcal{L}_\mathit{cnn}$. Two interesting misaligned pairs along the translation direction $\mathcal{L}_\mathit{comedy} \rightarrow \mathcal{L}_\mathit{fox}$ include $\langle \texttt{orange}, \texttt{dotard} \rangle$ and $\langle \texttt{blue}, \texttt{red} \rangle$.  

\subsection{All Four News Networks}

\begin{table}[htb]
\centering
\scriptsize
\setlength{\extrarowheight}{2pt}
\begin{tabular}{cc|c|c|c|c|c|c|c|c|c|c|c|c|c|}
  & \multicolumn{1}{c}{} & \multicolumn{4}{c}{$\mathcal{L}_\mathit{target}$} \\
  & \multicolumn{1}{c}{} &
  \multicolumn{1}{c}{$\mathcal{L}_\mathit{cnn}$}
  &\multicolumn{1}{c}{$\mathcal{L}_\mathit{fox}$}  & \multicolumn{1}{c}{$\mathcal{L}_\mathit{msnbc}$}  & \multicolumn{1}{c}{$\mathcal{L}_\mathit{oann}$} 
  \\\cline{3-6}
          \multirow{4}{*}{$\mathcal{L}_\mathit{source}$}    & $\mathcal{L}_\mathit{cnn}$ & \cellcolor{blue!25}- & 61.1\% & 62.0\% & 42.2\%  \\ \cline{3-6}
            & $\mathcal{L}_\mathit{fox}$ & 60.1\% & \cellcolor{blue!25}- & 53.2\% & 52.7\%  \\ \cline{3-6}
            & $\mathcal{L}_\mathit{msnbc}$ & 63.0\% & 52.8\% & \cellcolor{blue!25}- & 41.9\%  \\ \cline{3-6}
            & $\mathcal{L}_\mathit{oann}$ & 43.3\% & 54.8\% & 42.5\% & \cellcolor{blue!25}-  \\ \cline{3-6}
     
\end{tabular}
\vspace{0.2cm}
\caption{{Pairwise similarity between languages computed for the year 2020. Each corpus has identical number of tokens. The evaluation set (5K words) is computed by concatenating all three corpora and taking the top 5K words ranked by frequency.}}
\label{tab:oan}
\end{table}

Data set size is one of the most important contributing factors to ensure the quality of word embedding~\cite{NIPS2013_5021}. A large data set presents a word in richer contexts, ensuring that the embedding captures more semantic information. As shown in Figure~\ref{fig:newEngagement}, of all the four channels we consider, OANN has the least user engagement in terms of comments. After adding OANN in our comparison framework and sub-sampling all other corpora to match with $\mathcal{D}_\mathit{oann}$'s size, we observe that the pairwise similarity between all channels reduced. However, if we do not consider $\mathcal{L}_\mathit{oann}$ and just focus on the three languages,  the qualitative conclusions: (1) $\mathcal{L}_\mathit{cnn}$ is closer to $\mathcal{L}_\mathit{msnbc}$  than $\mathcal{L}_\mathit{fox}$; (2) $\mathcal{L}_\mathit{fox}$  is farthest from $\mathcal{L}_\mathit{msnbc}$; and (3) $\mathcal{L}_\mathit{msnbc}$  is closer to $\mathcal{L}_\mathit{cnn}$  and farthest from $\mathcal{L}_\mathit{fox}$, remain unaffected.

As shown in Table~\ref{tab:oan}, $\mathcal{L}_\mathit{oann}$ is farthest from $\mathcal{L}_\mathit{msnbc}$ and closest to $\mathcal{L}_\mathit{fox}$. However, $\mathcal{L}_\mathit{fox}$, a well-known conservative outlet, is closer to other two mainstream media outlets than $\mathcal{L}_\mathit{oann}$. In fact, a notable misaligned pair along the translation direction is $\mathcal{L}_\mathit{fox} \rightarrow \mathcal{L}_\mathit{oann}$ is $\langle \texttt{mask}, \texttt{muzzle} \rangle$.  We further note that if we arrange all channels along a political spectrum, a consistent arrangement is the following: MSNBC, CNN, Fox and finally, OANN (from left to right).

\subsection{Translating Trigrams}

Similar to the translation retrieval task presented in~\cite{SmithTHH17}, we conduct a translation retrieval task focused on high-frequency trigrams. Consistent with our earlier design choices, our source and target phrase vocabulary consist of 5,000 and 10,000 high-frequency trigrams of the combined 2020 Fox News and CNN corpora, respectively. Of the misaligned pairs we observe, the most notable is $\langle \texttt{black lives matter}, \texttt{all lives matter}\rangle$, \texttt{black lives matter} $\in \mathcal{L}_\mathit{cnn}$ and \texttt{all lives matter} $\in \mathcal{L}_\mathit{fox}$. 

\section{Conclusions and Future Work}

In this paper, we provide a novel perspective on analyzing political polarization through the lens of machine translation. Our simple-yet-powerful approach allows us to both gather statistical aggregates about large-scale discussion data sets, and at the same time zoom into nuanced differences in view points at the level of specific word pairs. Future lines of research include: (1) looking into the possibility of dialects (e.g., languages spoken by centrist Democrats and their more liberal counterpart); (2) further leverage unsupervised machine translation to perform political view translation at the level of phrases and sentences; and (3) robustness analysis on other social media platforms and countries.

\bibliographystyle{unsrt} 


\newpage
\section{Appendix}

\subsection{Experimental Setup}
Experiments are conducted in a suite of machines with the following specifications: 
\begin{compactitem}
\item OS: Windows 10.
\item Processor	Intel(R) Core(TM) i7-9750H CPU @ 2.60GHz, 2592 Mhz, 6 Core(s), 12 Logical Processor(s).
\item RAM: 64 GB.
\end{compactitem}

\subsection{Preprocessing and Hyperparameters}

To train word embedding on our data set, we use the following preprocessing steps. First, we remove all the emojis and non-ascii characters. Then, we remove all non-alphanumeric characters and lowercase the remaining text. We preserve the newline character after each individual document in the data set. We use the default parameters for training our FastText~\cite{bojanowski2017enriching} Skip-gram embedding with the dimension set to 100. 

\begin{table*}[htb]

{
\scriptsize
\begin{center}
     \begin{tabular}{| p{1.7cm}  |  p{5.8cm} |}
    \hline
    \Tstrut Category  & Misaligned pairs \\
     \hline 
 Political entities & 
 $\langle \texttt{democrats}, \texttt{republicans} \rangle$, $\langle \texttt{blue}, \texttt{red} \rangle$, $\langle \texttt{dem}, \texttt{republican} \rangle$, 
 $\langle \texttt{gop},	\texttt{democrats}\rangle$,
 $\langle \texttt{schumer},	\texttt{mcconnell}\rangle$\\
    \hline 

 News entities &  $\langle \texttt{fox}, \texttt{cnn} \rangle$, $\langle \texttt{tapper}, \texttt{carlson} \rangle$, $\langle \texttt{tapper}, \texttt{hannity} \rangle$, $\langle \texttt{tucker}, \texttt{cuomo} \rangle$, $\langle \texttt{lemon}, \texttt{hannity} \rangle$       \\
    \hline 
    	
 Derogatory & 
 $\langle \texttt{boarder}, \texttt{border} \rangle$,~\cite{Boarder}, $\langle \texttt{republicunts}, \texttt{democraps} \rangle$, 
 $\langle \texttt{maddow}, \texttt{madcow} \rangle$, 
 $\langle \texttt{democrats}, \texttt{demoncrats} \rangle$,  $\langle \texttt{cuomo}, \texttt{shithead} \rangle$, 
 $\langle \texttt{obama}, \texttt{obummer} \rangle$,  $\langle \texttt{schiff}, \texttt{schitt} \rangle$, 
 $\langle \texttt{spanky}, \texttt{trump} \rangle$
     \\
    \hline
(Near) synonyms & 
$\langle \texttt{lmao}, \texttt{lol} \rangle$, 
$\langle \texttt{stupidest}, \texttt{dumbest} \rangle$, $\langle \texttt{wh}, \texttt{whitehouse} \rangle$, 
$\langle \texttt{sociopath}, \texttt{psychopath} \rangle$, 
$\langle \texttt{favor}, \texttt{favour} \rangle$,  
$\langle \texttt{hahaha}, \texttt{hahahah} \rangle$,
$\langle \texttt{hillary}, \texttt{hrc} \rangle$, $\langle \texttt{congresswoman}, \texttt{pocahontas} \rangle$ 
	

\\
    \hline

Spelling errors & 
$\langle \texttt{melanie}, \texttt{melania} \rangle$, $\langle \texttt{kellyann}, \texttt{kellyanne} \rangle$,  
$\langle \texttt{hillary}, \texttt{hilary} \rangle$, 
$\langle \texttt{avenatti}, \texttt{avenati} \rangle$ 
     \\
    \hline

Ideological & 
$\langle \texttt{protests}, \texttt{riots} \rangle$, $\langle \texttt{progressives}, \texttt{socialists} \rangle$, 
$\langle \texttt{socialists}, \texttt{communists} \rangle$, $\langle \texttt{bigotry}, \texttt{paranoia} \rangle$, 
$\langle \texttt{liberals}, \texttt{conservatives} \rangle$, $\langle \texttt{communism}, \texttt{nazism} \rangle$, 
$\langle \texttt{commies}, \texttt{fascists} \rangle$, 
$\langle \texttt{liberalism}, \texttt{conservatism} \rangle$, $\langle \texttt{racism}, \texttt{supremacy} \rangle$    \\
    \hline

    \end{tabular}
\end{center}
\caption{{Characterizing the misaligned word pairs. We consider Fox News and CNN user discussions for the years 2017, 2018, 2019, and 2020.  Word pairs are presented in $\langle w_\mathit{cnn}, w_\mathit{fox} \rangle$ format where $w_\mathit{cnn} \in \mathcal{L}_\mathit{cnn}$ and $w_\mathit{fox} \in \mathcal{L}_\mathit{fox}$. Our findings are qualitatively similar to misaligned pairs listed in Table~\ref{tab:pairs}.}}
\label{tab:pairsUndedidated}}
\end{table*}

\begin{figure*}[htb]
\centering
\includegraphics[trim={0 0 0 0},clip, height=3.6in]{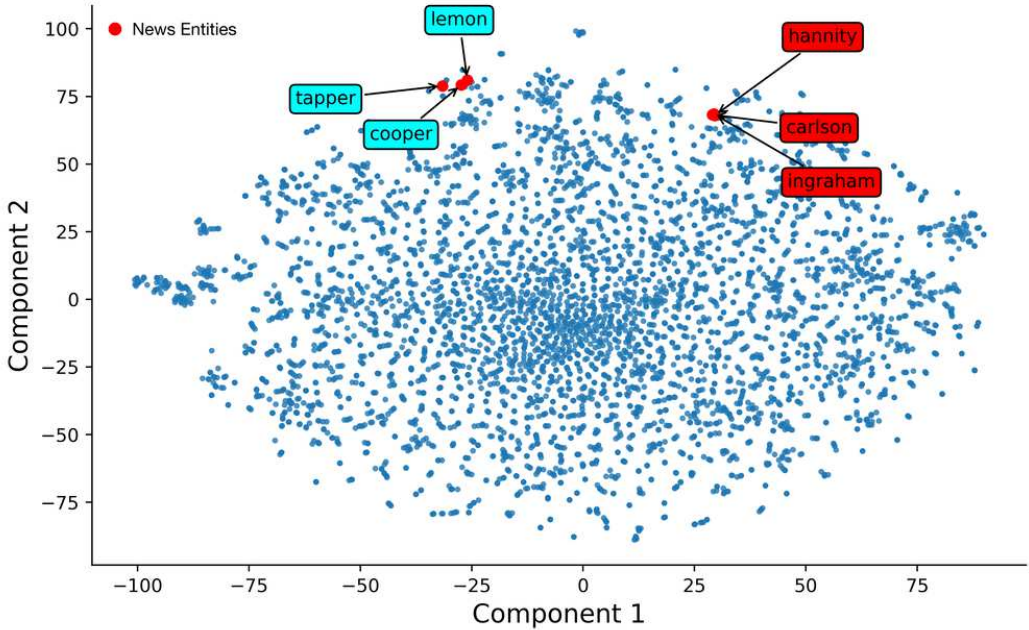}
\caption{A 2D t-SNE visualization~\cite{maaten2008visualizing} of top 1,000 tokens (shown in blue) and six news entities belonging to CNN and Fox News (shown in red) in CNN user discussions for the year 2020. We notice that three prominent news entities belonging to Fox News (Laura Ingraham, Sean Hannity, and Tucker Carlson) and CNN (Anderson Cooper, Jake Tapper, and Don Lemon) are clustered at two different locations. Illustrative examples highlighting that the misaligned pair $\langle \texttt{cooper}, \texttt{hannity}\rangle$ detected by our method appears in almost mirroring contexts in CNN and Fox News's users discussions are presented in Table~\ref{tab:hannitycooper}.} 
\label{fig:cooperHannity}
\end{figure*}

\begin{figure*}[htb]
\centering
\includegraphics[trim={0 0 0 0},clip, height=3.4in]{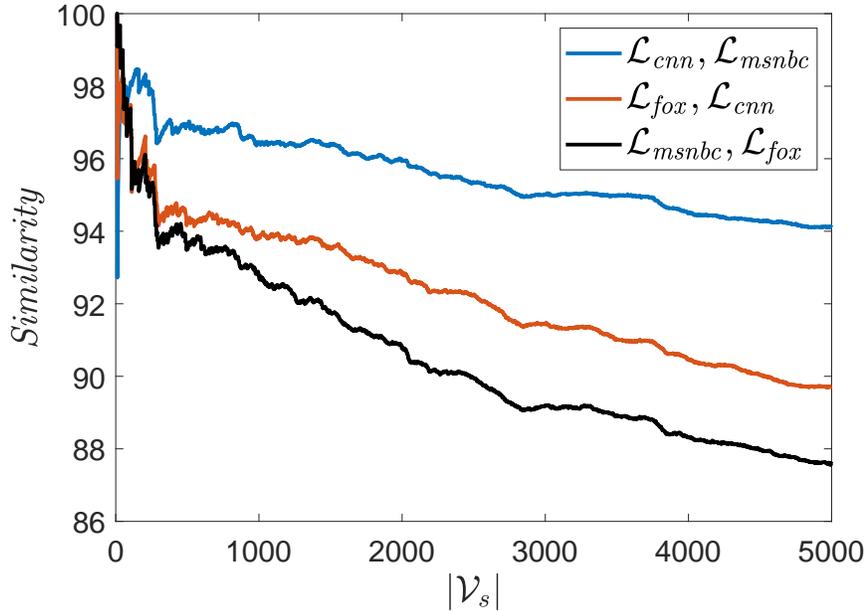}
\caption{$\mathit{Similarity}$ between language pairs as a function of the size of the source vocabulary $|\mathcal{V}_s|$ computed for the year 2019. For a given language pair $\mathcal{L}_i, \mathcal{L}_j$, we average five runs of  $\mathcal{L}_i \rightarrow \mathcal{L}_j$ and fives runs of $\mathcal{L}_j \rightarrow \mathcal{L}_i$.} 
\label{fig:parameter}
\end{figure*}

\begin{table*}[htb]

{
\scriptsize
\begin{center}
     \begin{tabular}{| p{7.5cm}  | p{7.5cm} |}
    \hline
    \Tstrut \cellcolor{blue!25} Cooper I love your show &\cellcolor{red!25} Hannity I love your show \\
     \hline \Tstrut
Mr.  \textcolor{blue}{Cooper I love your show}, but the more I watch your interview with this uneducated idiot, the more my stomach turns.  
 \Bstrut
 &      \textcolor{red}{Hannity I love your show}  I think you're very smart and it's clear you have great sources\ldots\\ 
 \hline
    \Tstrut 
    \cellcolor{blue!25} Cooper says it like it is &\cellcolor{red!25} Hannity says it like it is \\
    
    \hline
    \Tstrut
\textcolor{blue}{Cooper says it like it is}  it's not  fake news  it's  fake president
 \Bstrut
 & \textcolor{red}{Hannity says it like it is}\\ 
 \hline
    \Tstrut 
    \cellcolor{blue!25} Cooper keep up the good work &\cellcolor{red!25} Hannity keep up the good work  \\
     \hline \Tstrut
\textcolor{blue}{Cooper keep up the good work} and tell the American people the truth\ldots  \Bstrut
 &  \textcolor{red}{Hannity keep up the good work}. The truth is so refreshing and then we know how to vote\ldots\\
 \hline

    \end{tabular}
\end{center}
\caption{Illustrative examples highlighting that the discovered misaligned pair $\langle \texttt{cooper}, \texttt{hannity} \rangle $ are used in almost mirroring contexts. Left and right column contain user comments obtained from CNN and Fox news, respectively. }
\label{tab:hannitycooper}}
\end{table*}

\subsection{News Entities}

Figure~\ref{fig:cooperHannity} presents a 2D visualization of the word embedding space of CNN user discussions for the year 2020. We notice that three prominent news entities belonging to Fox News (Laura Ingraham, Sean Hannity, and Tucker Carlson) and CNN (Anderson Cooper, Jake Tapper, and Don Lemon) are clustered at two different locations. We further present example comments highlighting mirroring contexts for Sean Hannity and Anderson Cooper in Table~\ref{tab:hannitycooper}.

\subsection{Combining Replies}
Intuitively, in a politically tense environment, replies to a comment may contain views opposing the views presented in the comment (verified upon manual inspection). Hence, combining replies with comments may influence an increase in overall similarity between news networks' discussions. 
As shown in Table~\ref{tab:Replies} and Table~\ref{tab:2019}, our qualitative claim still holds, however, the similarities between news networks have increased. 

\begin{table}[htb]
\centering
\scriptsize
\setlength{\extrarowheight}{2pt}
\begin{tabular}{cc|c|c|c|}
  & \multicolumn{1}{c}{} & \multicolumn{3}{c}{$\mathcal{L}_{\emph{target}}$} \\
  & \multicolumn{1}{c}{} & \multicolumn{1}{c}{$\mathcal{L}_{\emph{cnn}}$}  & \multicolumn{1}{c}{$\mathcal{L}_{\emph{fox}}$}  & \multicolumn{1}{c}{$\mathcal{L}_{\emph{msnbc}}$} \\\cline{3-5}
            & $\mathcal{L}_{\emph{cnn}}$ &\cellcolor{blue!25} - & \textcolor{black}{92.5\%}  & \textcolor{black}{94.0\%}
 \\ \cline{3-5}
$\mathcal{L}_{\emph{source}}$  & $\mathcal{L}_{\emph{fox}}$ & \textcolor{black}{92.3\%} &\cellcolor{blue!25} - & \textcolor{black}{90.4\%}
 \\\cline{3-5}
            & $\mathcal{L}_{\emph{msnbc}}$ & \textcolor{black}{93.9\%} & \textcolor{black}{90.7\%}  &\cellcolor{blue!25} - \\\cline{3-5}
\end{tabular}
\vspace{0.2cm}
\caption{{Pairwise similarity between languages computed for the year 2019 combining comments and replies. Each corpus has identical number of tokens. The evaluation set (5K words) is computed by concatenating all three corpora and taking the top 5K words ranked by frequency. Since stopwords are used as anchor words, stopwords are excluded in the evaluation set.}}
\label{tab:Replies}
\end{table}

\subsection{Additional Years}

We present the results computed for the year 2018 and 2019 in Table~\ref{tab:2018} and \ref{tab:2019}, respectively. We notice that our initial observation reported in Table~\ref{tab:threeChannels} that Fox News and MSNBC are the farthest apart with CNN placed in the middle, holds in our new experiments summarized in Table~\ref{tab:2018} and \ref{tab:2019}. 

\begin{table*}[htb]
\centering
\scriptsize
\setlength{\extrarowheight}{2pt}
\begin{tabular}{cc|c|c|c|}
  & \multicolumn{1}{c}{} & \multicolumn{3}{c}{$\mathcal{L}_{\emph{target}}$} \\
  & \multicolumn{1}{c}{} & \multicolumn{1}{c}{$\mathcal{L}_{\emph{cnn}}$}  & \multicolumn{1}{c}{$\mathcal{L}_{\emph{fox}}$}  & \multicolumn{1}{c}{$\mathcal{L}_{\emph{msnbc}}$} \\\cline{3-5}
            & $\mathcal{L}_{\emph{cnn}}$ &\cellcolor{blue!25} - & \textcolor{black}{87.7\%}  & \textcolor{black}{86.1\%}
 \\ \cline{3-5}
$\mathcal{L}_{\emph{source}}$  & $\mathcal{L}_{\emph{fox}}$ & \textcolor{black}{88.7\%} &\cellcolor{blue!25} - & \textcolor{black}{78.3\%}
 \\\cline{3-5}
            & $\mathcal{L}_{\emph{msnbc}}$ & \textcolor{black}{87.0\%} & \textcolor{black}{79.3\%}  &\cellcolor{blue!25} - \\\cline{3-5}
\end{tabular}
\vspace{0.5cm}
\caption{{Pairwise similarity between languages computed for the year 2018. Each corpus has identical number of tokens. The evaluation set (5K words) is computed by concatenating all three corpora and taking the top 5K words ranked by frequency. Since stopwords are used as anchor words, stopwords are excluded in the evaluation set.}}
\label{tab:2018}
\end{table*}

\begin{table*}[htb]
\centering
\scriptsize
\setlength{\extrarowheight}{2pt}
\begin{tabular}{cc|c|c|c|}
  & \multicolumn{1}{c}{} & \multicolumn{3}{c}{$\mathcal{L}_{\emph{target}}$} \\
  & \multicolumn{1}{c}{} & \multicolumn{1}{c}{$\mathcal{L}_{\emph{cnn}}$}  & \multicolumn{1}{c}{$\mathcal{L}_{\emph{fox}}$}  & \multicolumn{1}{c}{$\mathcal{L}_{\emph{msnbc}}$} \\\cline{3-5}
            & $\mathcal{L}_{\emph{cnn}}$ &\cellcolor{blue!25} - & \textcolor{black}{89.3\%}  & \textcolor{black}{90.6\%}
 \\ \cline{3-5}
$\mathcal{L}_{\emph{source}}$  & $\mathcal{L}_{\emph{fox}}$ & \textcolor{black}{89.2\%} &\cellcolor{blue!25} - & \textcolor{black}{86.1\%}
 \\\cline{3-5}
            & $\mathcal{L}_{\emph{msnbc}}$ & \textcolor{black}{90.8\%} & \textcolor{black}{86.3\%}  &\cellcolor{blue!25} - \\\cline{3-5}
\end{tabular}
\vspace{0.2cm}
\caption{{Pairwise similarity between languages computed for the year 2019. Each corpus has identical number of tokens. The evaluation set (5K words) is computed by concatenating all three corpora and taking the top 5K words ranked by frequency. Since stopwords are used as anchor words, stopwords are excluded in the evaluation set.}}
\label{tab:2019}
\end{table*}

\subsection{User Filter}

Our user filter assigns one user to a specific news network. Intuitively, if there are one conservative news network and two liberal news networks, our filter may cause a split among the liberal users causing some minor shifts of the network in the center toward the right. We observe this phenomenon in the difference between Table~\ref{tab:threeChannels} (which does not use our filter) and Table~\ref{tab:Filter} (which uses our filter). We still find that our initial arrangement obtained in Table~\ref{tab:threeChannels} -- MSNBC, CNN and Fox News (from left to right) -- is consistent with the newly obtained results in Table~\ref{tab:Filter}. However, we do notice that in Table~\ref{tab:Filter}, $\mathcal{L}_\mathit{cnn}$ gets closer to $\mathcal{L}_\mathit{fox}$ than $\mathcal{L}_\mathit{msnbc}$. Qualitatively similar to Table~\ref{tab:pairs} (obtained using our filter), we obtain several misaligned pairs listed in Table~\ref{tab:pairsUndedidated} (obtained without using our filter).


\begin{table*}[htb]
\centering
\scriptsize
\setlength{\extrarowheight}{2pt}
\begin{tabular}{cc|c|c|c|}
  & \multicolumn{1}{c}{} & \multicolumn{3}{c}{$\mathcal{L}_{\emph{target}}$} \\
  & \multicolumn{1}{c}{} & \multicolumn{1}{c}{$\mathcal{L}_{\emph{cnn}}$}  & \multicolumn{1}{c}{$\mathcal{L}_{\emph{fox}}$}  & \multicolumn{1}{c}{$\mathcal{L}_{\emph{msnbc}}$} \\\cline{3-5}
            & $\mathcal{L}_{\emph{cnn}}$ &\cellcolor{blue!25} - & \textcolor{black}{84.3\%}  & \textcolor{black}{82.0\%}
 \\ \cline{3-5}
$\mathcal{L}_{\emph{source}}$  & $\mathcal{L}_{\emph{fox}}$ & \textcolor{black}{84.6\%} &\cellcolor{blue!25} - & \textcolor{black}{77.2\%}
 \\\cline{3-5}
            & $\mathcal{L}_{\emph{msnbc}}$ & \textcolor{black}{82.9\%} & \textcolor{black}{79.3\%}  &\cellcolor{blue!25} - \\\cline{3-5}
\end{tabular}
\vspace{0.2cm}
\caption{{Pairwise similarity between languages computed for the year 2020 when user filter is used. Each corpus has identical number of tokens. The evaluation set (5K words) is computed by concatenating all three corpora and taking the top 5K words ranked by frequency. Since stopwords are used as anchor words, stopwords are excluded in the evaluation set.}}
\label{tab:Filter}
\end{table*}

\subsection{Comedy Data Set}

Overall, we considered 3,898 videos obtained from the official YouTube channels of the comedians listed in Table~\ref{tab:channelsComedy}. 4,099,081 comments are obtained from these videos using the publicly available YouTube API. 

\begin{table*}[htb]
{
\begin{center}
     
\begin{tabular}{|l | c | c |}
    \hline
    Comedian & \#Subscribers & \#Videos \\
    \hline                                 
   Trevor Noah  &  8.3M & 729 \\ 
    \hline
   Stephen Colbert & 7.8M & 1,428 \\
    \hline
   Jimmy Kimmel  &  17.0M & 935 \\
    \hline
   John Oliver  &  8.4M & 35 \\
    \hline
   Seth Meyers  &  3.8M & 771 \\
    \hline
    \end{tabular}

\caption{List of YouTube channels of comedians considered in results reported in Table~\ref{tab:comedy}. Video count reflects \#videos uploaded on or before 31 Dec 2019 starting from 1 January 2019.}
\label{tab:channelsComedy}
\end{center}
}
\end{table*}

\subsection{Robustness Analysis}

\subsubsection{Other machine translation algorithms} As shown in Table~\ref{tab:Lample} and Table~\ref{tab:2019}, we obtain consistent results with a different machine translation algorithm~\cite{LampleCRDJ18}. The overall conclusion of Fox News and MSNBC being farthest apart, remains unchanged. 


\begin{table*}[htb]
\centering
\scriptsize
\setlength{\extrarowheight}{2pt}
\begin{tabular}{cc|c|c|c|}
  & \multicolumn{1}{c}{} & \multicolumn{3}{c}{$\mathcal{L}_{\emph{target}}$} \\
  & \multicolumn{1}{c}{} & \multicolumn{1}{c}{$\mathcal{L}_{\emph{cnn}}$}  & \multicolumn{1}{c}{$\mathcal{L}_{\emph{fox}}$}  & \multicolumn{1}{c}{$\mathcal{L}_{\emph{msnbc}}$} \\\cline{3-5}
            & $\mathcal{L}_{\emph{cnn}}$ &\cellcolor{blue!25} - & \textcolor{black}{86.9\%}  & \textcolor{black}{87.6\%}
 \\ \cline{3-5}
$\mathcal{L}_{\emph{source}}$  & $\mathcal{L}_{\emph{fox}}$ & \textcolor{black}{86.5\%} &\cellcolor{blue!25} - & \textcolor{black}{81.8\%}
 \\\cline{3-5}
            & $\mathcal{L}_{\emph{msnbc}}$ & \textcolor{black}{88.0\%} & \textcolor{black}{82.2\%}  &\cellcolor{blue!25} - \\\cline{3-5}
\end{tabular}
\vspace{0.2cm}
\caption{{Pairwise similarity between languages computed for the year 2019 using~\cite{LampleCRDJ18}. Each corpus has identical number of tokens. The evaluation set (5K words) is computed by concatenating all three corpora and taking the top 5K words ranked by frequency. Instead of algorithm proposed by~\cite{SmithTHH17}, we present results using a different machine translation algorithm proposed by~\cite{LampleCRDJ18}. While using this new translation algorithm, we use the same set of stopwords as anchor words.}}
\label{tab:Lample}
\end{table*}

\begin{table*}[htb]
\centering
\scriptsize
\setlength{\extrarowheight}{2pt}
\begin{tabular}{cc|c|c|c|}
  & \multicolumn{1}{c}{} & \multicolumn{3}{c}{$\mathcal{L}_{\emph{target}}$} \\
  & \multicolumn{1}{c}{} & \multicolumn{1}{c}{$\mathcal{L}_{\emph{cnn}}$}  & \multicolumn{1}{c}{$\mathcal{L}_{\emph{fox}}$}  & \multicolumn{1}{c}{$\mathcal{L}_{\emph{msnbc}}$} \\\cline{3-5}
            & $\mathcal{L}_{\emph{cnn}}$ &\cellcolor{blue!25} - & \textcolor{black}{37.9 $\pm$ 0.75\%}  & \textcolor{black}{41.5 $\pm$ 0.59\%}
 \\ \cline{3-5}
$\mathcal{L}_{\emph{source}}$  & $\mathcal{L}_{\emph{fox}}$ & \textcolor{black}{37.9 $\pm$ 0.73\%} &\cellcolor{blue!25} - & \textcolor{black}{35.6 $\pm$ 0.43\%}
 \\\cline{3-5}
            & $\mathcal{L}_{\emph{msnbc}}$ & \textcolor{black}{41.5 $\pm$ 0.55\%} & \textcolor{black}{35.4 $\pm$ 0.24\%}  &\cellcolor{blue!25} - \\\cline{3-5}
\end{tabular}
\vspace{0.2cm}
\caption{{Pairwise similarity between languages computed for the year 2020 using our new similarity measure $\mathit{Similarity}_\mathcal{N}$. Each corpus has identical number of tokens. The evaluation set (5K words) is computed by concatenating all three corpora and taking the top 5K words ranked by frequency. Since stopwords are used as anchor words, stopwords are excluded in the evaluation set. Each cell summarizes the mean and standard deviation of five runs. Our obtained results using this new measure
are consistent with the results presented in Table~\ref{tab:threeChannelsMultipleRuns}.}}
\label{tab:threeChannelsNeighbor}
\end{table*}

\begin{table*}[htb]
\centering
\scriptsize
\setlength{\extrarowheight}{2pt}
\begin{tabular}{cc|c|c|c|}
  & \multicolumn{1}{c}{} & \multicolumn{3}{c}{$\mathcal{L}_{\emph{target}}$} \\
  & \multicolumn{1}{c}{} & \multicolumn{1}{c}{$\mathcal{L}_{\emph{cnn}}$}  & \multicolumn{1}{c}{$\mathcal{L}_{\emph{fox}}$}  & \multicolumn{1}{c}{$\mathcal{L}_{\emph{msnbc}}$} \\\cline{3-5}
            & $\mathcal{L}_{\emph{cnn}}$ &\cellcolor{blue!25} - & \textcolor{black}{90.0 $\pm$ 1.05\%}  & \textcolor{black}{94.2 $\pm$ 0.30\%}
 \\ \cline{3-5}
$\mathcal{L}_{\emph{source}}$  & $\mathcal{L}_{\emph{fox}}$ & \textcolor{black}{89.4 $\pm$ 0.89\%} &\cellcolor{blue!25} - & \textcolor{black}{87.0 $\pm$ 0.52\%}
 \\\cline{3-5}
            & $\mathcal{L}_{\emph{msnbc}}$ & \textcolor{black}{94.0 $\pm$ 0.35\%} & \textcolor{black}{88.2 $\pm$ 0.35\%}  &\cellcolor{blue!25} - \\\cline{3-5}
\end{tabular}
\caption{{Pairwise similarity between languages computed for the year 2020. Each corpus has identical number of tokens. The evaluation set (5K words) is computed by concatenating all three corpora and taking the top 5K words ranked by frequency. Since stopwords are used as anchor words, stopwords are excluded in the evaluation set. Each cell summarizes the mean and standard deviation of five runs.}}
\label{tab:threeChannelsMultipleRuns}
\end{table*}

\subsubsection{Stability across multiple runs}
For each translation direction, we run five experiments with different splits of our data sets. As shown in Table~\ref{tab:threeChannelsMultipleRuns}, our similarity results are stable across multiple runs. Moreover, our prior arrangement of the news networks along a political spectrum obtained from the results presented in Table~\ref{tab:threeChannels} -- MSNBC, CNN and Fox News (from left to right) -- still holds.

\subsubsection{Choice of $|\mathcal{V}_\mathit{s}|$} In our experiments, $\mathcal{V}_s \subset \mathcal{V}_t$ and we select the top 5K words ranked by frequency from $\mathcal{V}_t$ as $\mathcal{V}_s$. Effectively, the size of $\mathcal{V}_s$ is a  hyperparameter set to 5K. As shown in Figure~\ref{fig:parameter}, beyond a reasonable threshold of 1,000, our results are not sensitive to the choice for this hyperparameter. Since we have observed that our similarity measure is reasonably symmetric, in this figure, we average 10 runs of a given language pair (five runs along $\mathcal{L}_i \rightarrow \mathcal{L}_j$  and five runs along $\mathcal{L}_j \rightarrow \mathcal{L}_i$). We demonstrate that our overall conclusion -- Fox News and MSNBC are farthest apart -- is reasonably robust to the choice of this hyperparameter.

\subsubsection{Fine-grained similarity measure} Recall that, our similarity measure between two languages computes the fraction of words in $\mathcal{V}_s$ that translates to themselves. It is possible that even though a source word translates to itself, the neighborhoods of the source and the target word in their respective embedding spaces may not be identical. In what follows, in order to capture this difference, we present a fine-grained similarity measure. Let $\mathcal{N}(w)^{k,l}$ denote the top $k$ neighbors of the word $w$ in an embedding space trained on $\mathcal{D}_l$. In our new measure, $\mathit{Similarity}_\mathcal{N}$, we define the similarity between two languages as follows:\\ {
\emph{Similarity}$_\mathcal{N}$($\mathcal{L_\mathit{s}}, \mathcal{L_\mathit{t}})$ = $\frac{\Sigma_{w_s \in \mathcal{V}_\mathit{s}} \mathit{Jaccard}(\mathcal{N}(w_s)^{k,s} ,~\mathcal{N}(w_t)^{k,t} )}{|\mathcal{V}_\mathit{s}|}$
} where $w_t = \emph{translate}(w_s)^{\mathcal{L}_\mathit{s} \rightarrow \mathcal{L}_\mathit{t}}$ and $\mathit{Jaccard}(A,B)$ takes two sets $A$ and $B$ as inputs and outputs the Jaccard similarity between the two sets (defined as $\frac{|A \cap B|}{|A \cup B|}\times100$ for sets $A$ and $B$). As shown in Table~\ref{tab:threeChannelsMultipleRuns} and \ref{tab:threeChannelsNeighbor}, our obtained results using our new measure in Table~\ref{tab:threeChannelsNeighbor} ($k$ is set to 10)
are consistent with the results presented in Table~\ref{tab:threeChannelsMultipleRuns}. 

\subsection{Evolving Trends in Comments Share}

Let $\mathit{u}_\mathit{cnn}$   denote the total number of comments a unique user $u$ posts on videos uploaded by CNN in a given year and $\mathit{u}_\mathit{fox}$   denote the total number of comments a unique user $u$ posts on videos uploaded by Fox news in a given year. We now define the following sets (the superscript indicating the news network where the comments are posted):\\ 
(1) $\mathit{cnn}_\mathit{sole}^\mathit{cnn}$ as comments posted by a user $u$ such that $\mathit{u}_\mathit{fox} = 0$ and $\mathit{u}_\mathit{cnn} > 0$;\\ 
(2) $\mathit{fox}_\mathit{sole}^\mathit{fox}$ as comments posted by a user $u$ such that $\mathit{u}_\mathit{cnn} = 0$ and $\mathit{u}_\mathit{fox} > 0$;\\
(3) $\mathit{cnn}_\mathit{maj}^\mathit{cnn}$ as CNN comments posted by a user $u$ such that $\mathit{u}_\mathit{cnn} > 0$, $\mathit{u}_\mathit{fox} > 0$, and 
$\mathit{u}_\mathit{cnn} > \mathit{u}_\mathit{fox}$;\\
(4) $\mathit{cnn}_\mathit{maj}^\mathit{fox}$ as Fox news comments posted by a user $u$ such that $\mathit{u}_\mathit{cnn} > 0$, $\mathit{u}_\mathit{fox} > 0$, and 
$\mathit{u}_\mathit{cnn} > \mathit{u}_\mathit{fox}$;\\
(5) $\mathit{fox}_\mathit{maj}^\mathit{cnn}$ as CNN comments posted by a user $u$ such that $\mathit{u}_\mathit{cnn} > 0$, $\mathit{u}_\mathit{fox} > 0$, and 
$\mathit{u}_\mathit{fox} > \mathit{u}_\mathit{cnn}$;\\
(6) $\mathit{fox}_\mathit{maj}^\mathit{fox}$ as Fox news comments posted by a user $u$ such that $\mathit{u}_\mathit{cnn} > 0$, $\mathit{u}_\mathit{fox} > 0$, and 
$\mathit{u}_\mathit{fox} > \mathit{u}_\mathit{cnn}$;\\
 and\\
(7) $\mathit{equal}$ as comments posted by a user $u$ such that $\mathit{u}_\mathit{cnn} > 0$, $\mathit{u}_\mathit{fox} > 0$, and 
$\mathit{u}_\mathit{cnn} = \mathit{u}_\mathit{fox}$.

\begin{figure*}[htb]

\centering
\subfigure[2020]{%
\includegraphics[width = 0.40 \textwidth]{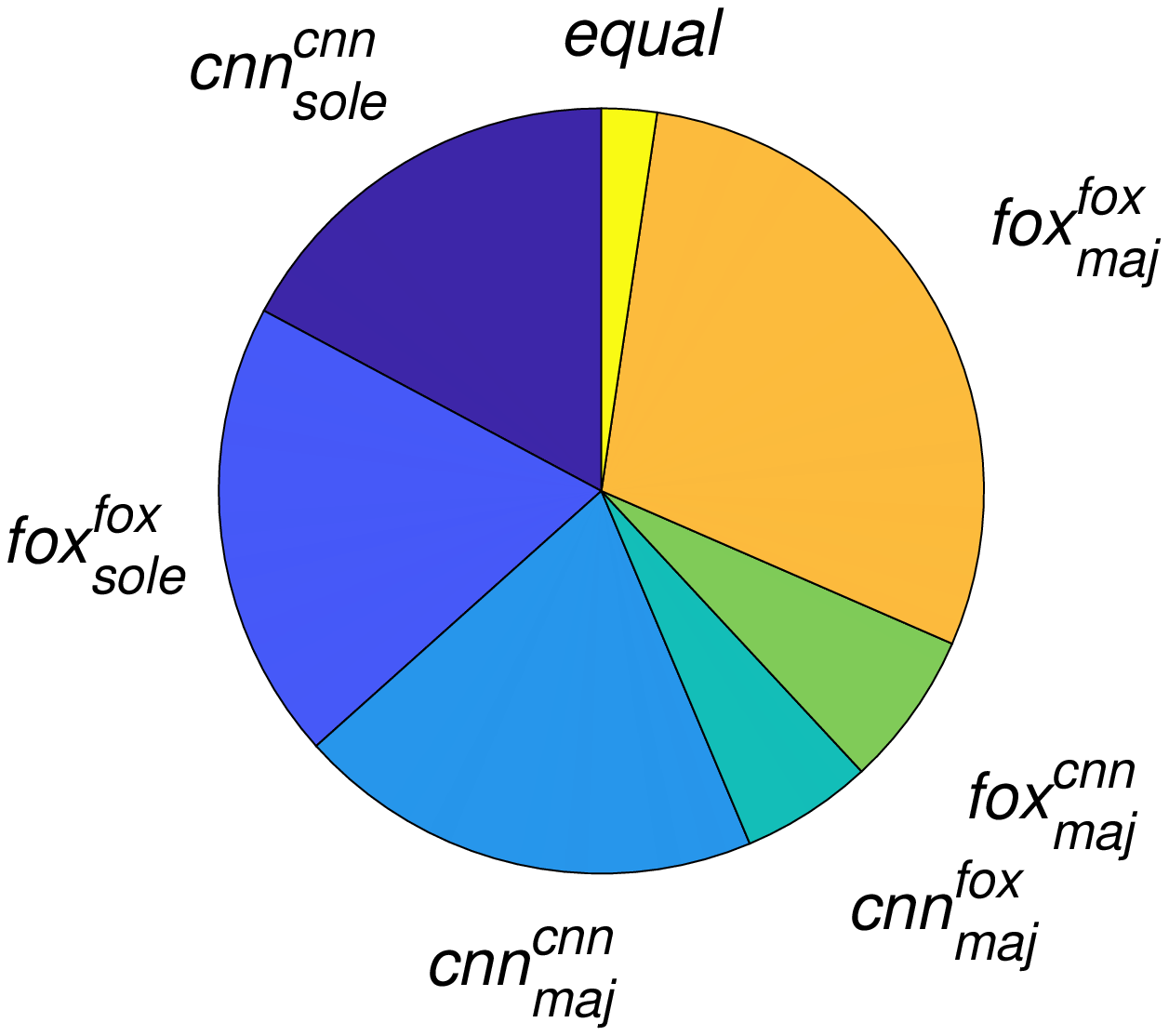}
\label{fig:2020}}
\subfigure[2019]{%
\includegraphics[width = 0.40 \textwidth]{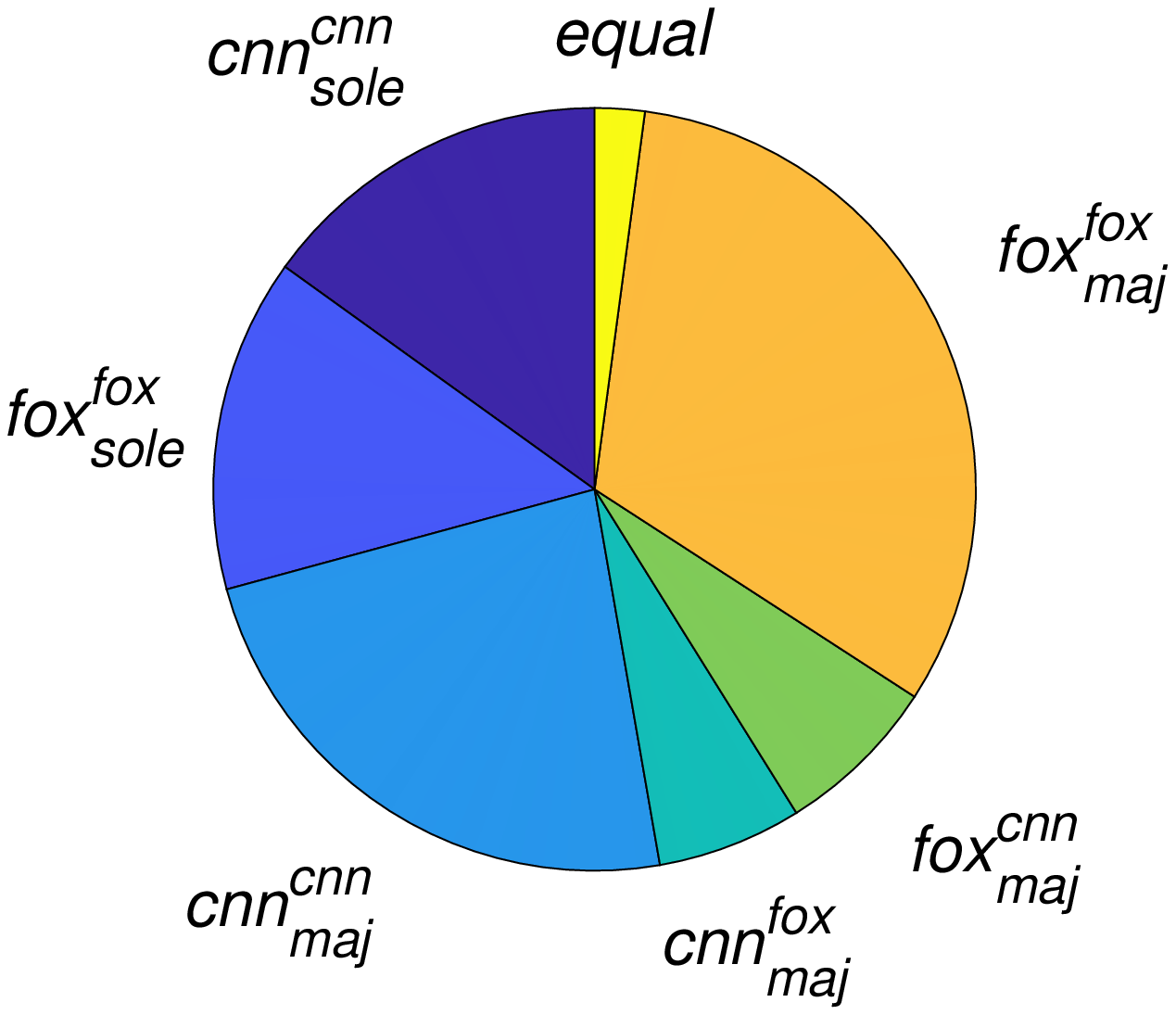}
\label{fig:2019}}
\subfigure[2018]{%
\includegraphics[width = 0.40 \textwidth]{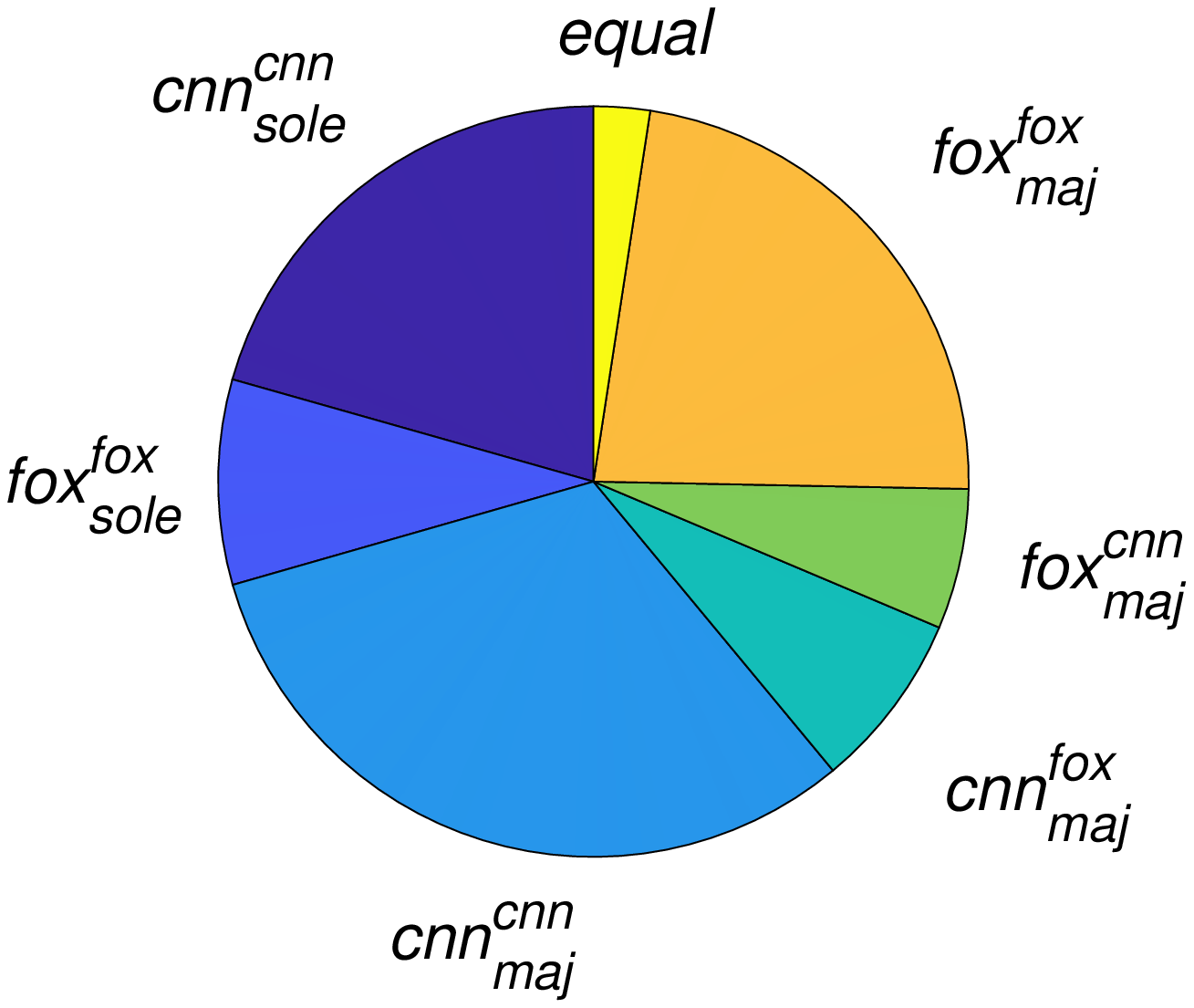}
\label{fig:2018}}
\subfigure[2017]{%
\includegraphics[width = 0.40 \textwidth]{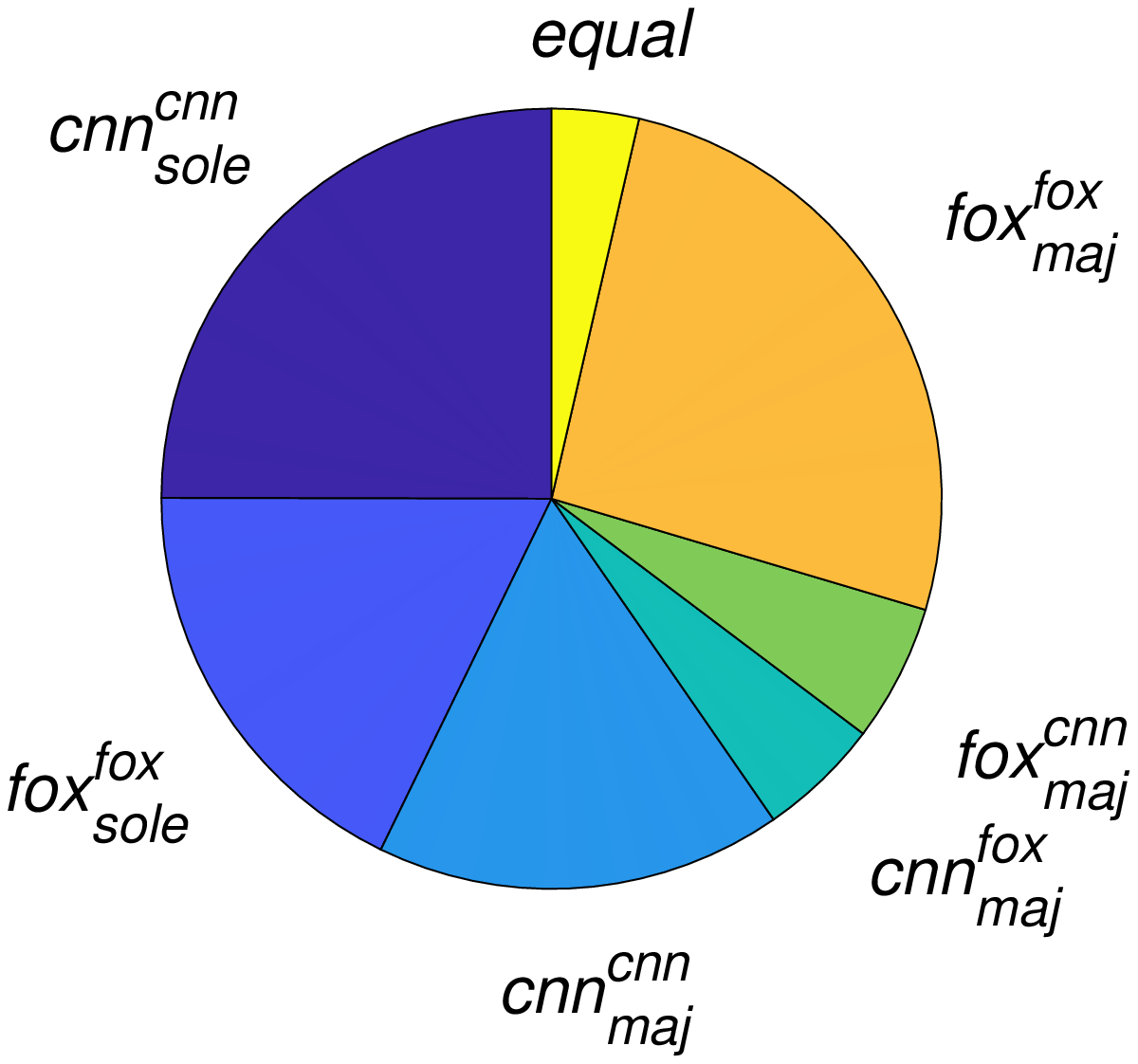}
\label{fig:2017}}
\subfigure[2016]{%
\includegraphics[width = 0.40 \textwidth]{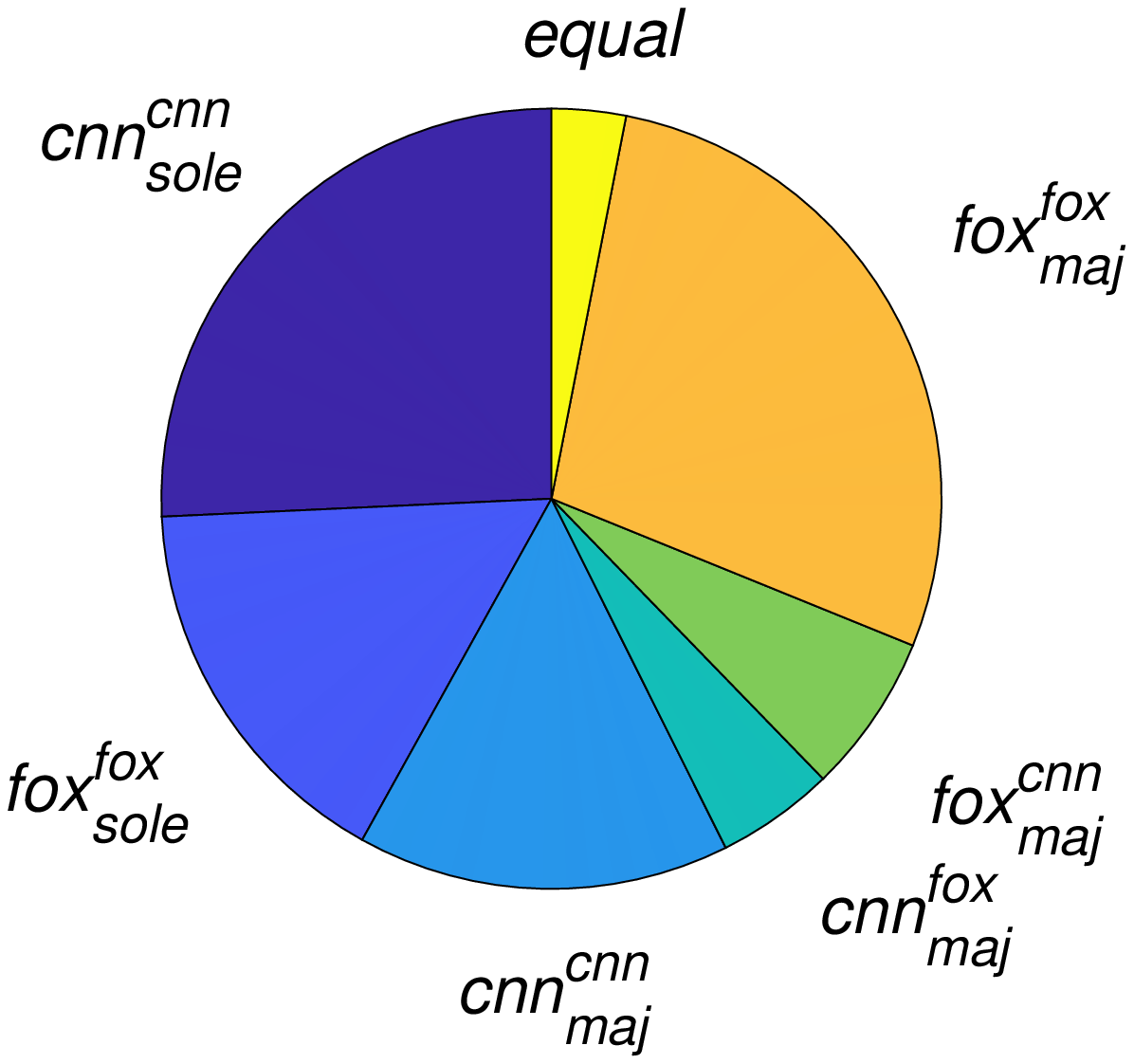}
\label{fig:2016}}
\subfigure[2015]{%
\includegraphics[width = 0.40 \textwidth]{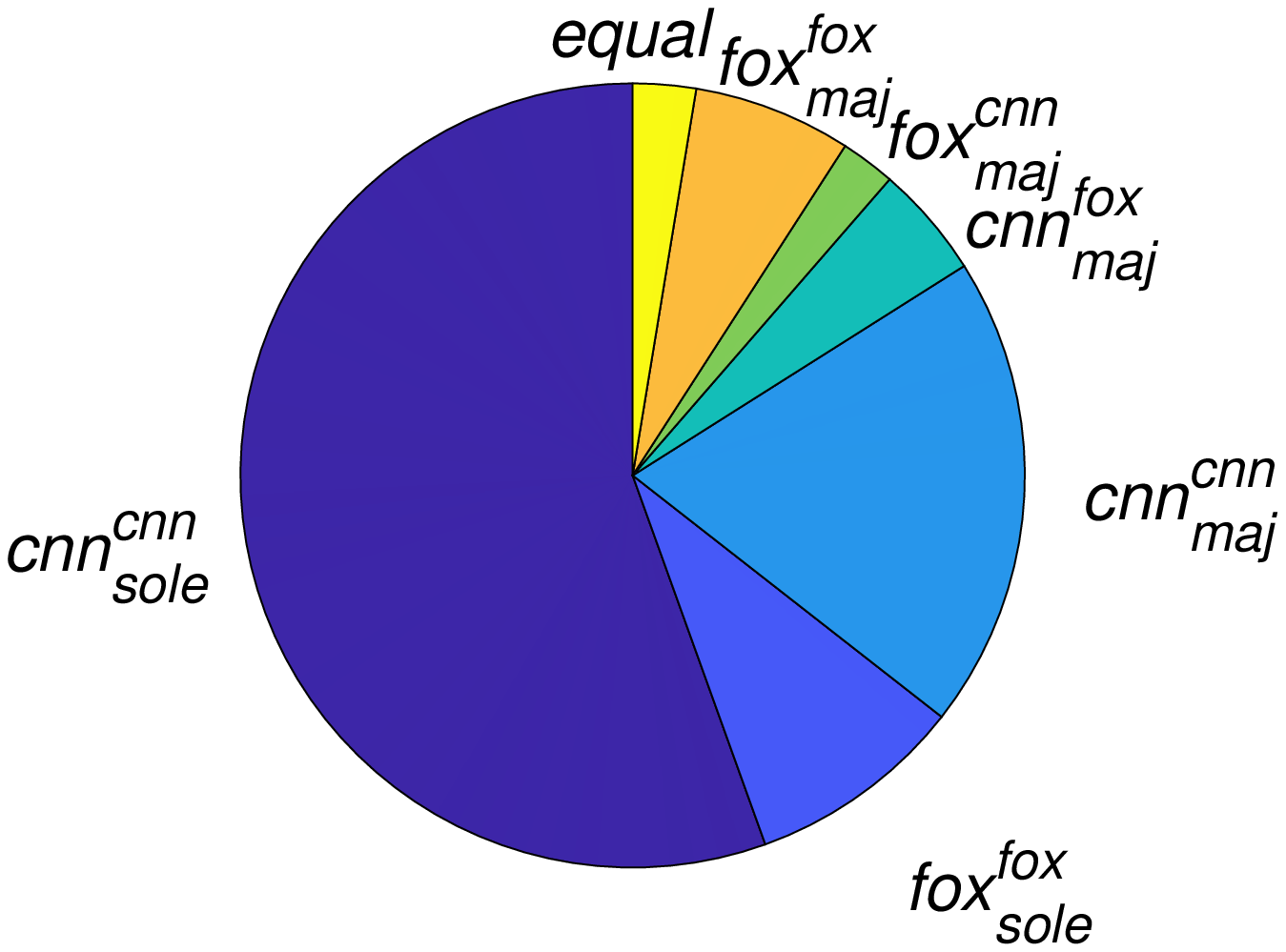}
\label{fig:2015}}

\caption{{Comment distributions over the years.}}

\label{fig:comment}
\end{figure*}

Figure~\ref{fig:comment} presents the relative distributions of the seven categories we defined above. We note that both $\mathit{fox}_\mathit{maj}^\mathit{cnn}$ and $\mathit{cnn}_\mathit{maj}^\mathit{fox}$, i.e., comments made by users on a news network who are frequent commenters of the other news network, almost mirrors each other. We also note that from a sparse presence in 2015, engagement in Fox News grew remarkably in the year 2016 and since then, the engagement is comparable with CNN.   

\end{document}